%% file: main.tex
\documentclass[sigconf,screen,nonacm]{acmart}

\setcopyright{none}
\acmPrice{15.00}
\acmDOI{10.1145/3468264.3468593}
\acmYear{2021}
\copyrightyear{2021}
\acmSubmissionID{fse21main-p521-p}
\acmISBN{978-1-4503-8562-6/21/08}
\acmConference[ESEC/FSE '21]{Proceedings of the 29th ACM Joint European Software Engineering Conference and Symposium on the Foundations of Software Engineering}{August 23--28, 2021}{Athens, Greece}
\acmBooktitle{Proceedings of the 29th ACM Joint European Software Engineering Conference and Symposium on the Foundations of Software Engineering (ESEC/FSE '21), August 23--28, 2021, Athens, Greece}

\usepackage[utf8x]{inputenc}
\usepackage{booktabs}   %

\usepackage{graphicx}
\usepackage{microtype} %

\usepackage{amsmath}
\usepackage{algorithm}
\usepackage{algpseudocode}
\usepackage{dsfont}
\usepackage{paralist}
\usepackage[noabbrev,capitalize]{cleveref}
\crefname{lstlisting}{Listing}{listings}
\usepackage{listings}
\definecolor{codegreen}{rgb}{0,0.6,0}
\definecolor{codegray}{rgb}{0.5,0.5,0.5}
\definecolor{codepurple}{rgb}{0.58,0,0.82}
\definecolor{backcolour}{rgb}{0.95,0.95,0.92}
\lstdefinestyle{mystyle}{
    commentstyle=\color{codegreen},
    keywordstyle=\color{magenta},
    numberstyle=\tiny\color{codegray},
    stringstyle=\color{codepurple},
    basicstyle=\ttfamily\footnotesize,
    breakatwhitespace=false,
    breaklines=true,
    captionpos=b,
    keepspaces=true,
    numbers=left,
    numbersep=5pt,
    showspaces=false,
    frame=tblr,
    showstringspaces=false,
    showtabs=false,
    xleftmargin=10pt,
    framexleftmargin=10pt,
    framexrightmargin=0pt,
    framexbottommargin=4pt,
    tabsize=2
}

\newcommand{\cmt}[2]{}

\newcommand{\af}[1]{\cmt{\color{blue}{AF}}{\color{blue} {#1}}}

\input{defns}

\fussy
\begin{document}
\title{Symbolic Parallel Adaptive Importance Sampling for Probabilistic Program Analysis}

\author{Yicheng Luo}
\authornote{Work done when the author is with Imperial College London}
\orcid{0000-0003-0547-411X}
\affiliation{%
  \institution{University College London}
  \institution{Imperial College London}
  \country{United Kingdom}
}
\email{yicheng.luo.20@ucl.ac.uk}

\author{Antonio Filieri}
\orcid{0000-0001-9646-646X}
\affiliation{%
  \institution{Imperial College London}
  \country{United Kingdom}
}
\email{a.filieri@imperial.ac.uk}

\author{Yuan Zhou}
\orcid{0000-0002-3103-5535}
\affiliation{%
  \institution{Artificial Intelligence \\ Research Center, DII}
  \country{China}
}
\email{yuaanzhou@outlook.com}

\begin{abstract}
Probabilistic software analysis aims at quantifying the probability of a target event
occurring during the execution of a program processing uncertain incoming data or
written itself using probabilistic programming constructs. Recent techniques combine
symbolic execution with model counting or solution space quantification methods to
obtain accurate estimates of the occurrence probability of rare target events, such as
failures in a mission-critical system. However, they face several scalability and
applicability limitations when analyzing software processing with high-dimensional and
correlated multivariate input distributions.

In this paper, we present SYMbolic Parallel Adaptive Importance Sampling (SYMPAIS), a
new inference method tailored to analyze path conditions generated from the symbolic
execution of programs with high-dimensional, correlated input distributions.
SYMPAIS combines results from importance sampling and constraint solving to produce
accurate estimates of the satisfaction probability for a broad class of constraints that
cannot be analyzed by current solution space quantification methods.
We demonstrate SYMPAIS's generality and performance compared with state-of-the-art
alternatives on a set of problems from different application domains.
\end{abstract}

\ccsdesc[500]{Mathematics of computing~Metropolis-Hastings algorithm}
\ccsdesc[500]{Software and its engineering~Software verification and validation}

\keywords{symbolic execution, probabilistic analysis, probabilistic programming,
importance sampling, Markov chain Monte Carlo}

\maketitle

\input{sections/introduction}
\input{sections/background}

\input{sections/sympais}
\input{sections/evaluation}
\input{sections/related_work}
\vspace{-2mm}
\input{sections/conclusions}
\begin{acks}                            %
  We would like to thank the anonymous reviewers for useful comments
  and feedback. Yicheng Luo is supported by \grantsponsor{UKRI}{UKRI AI Centre for
  Doctoral Training in Foundational Artificial
  Intelligence}{https://gow.epsrc.ukri.org/NGBOViewGrant.aspx?GrantRef=EP/S021566/1}
  (\grantnum{UKRI}{EP/S021566/1}) and the UCL Overseas Research Scholarship (UCL-ORS).
  Part of the experiments has been enabled by an equipment grant from the UK National Cyber Security Centre (NCSC) awarded to Antonio Filieri. 
  Yuan Zhou is partially supported by the \grantsponsor{NSFC}{National Natural Science
  Foundation of China (NSFC)}{http://www.nsfc.gov.cn}.
\end{acks}

\clearpage
\appendix\label{appendix}
\input{sections/appendix}

\newpage
\bibliography{references}

\end{document}

%% file: defns.tex
\newcommand{\myvec}[1]{\boldsymbol{#1}}
\newcommand{\myvecsym}[1]{\boldsymbol{#1}}
\newcommand{\ind}{\mathds{1}}

\newcommand{\vmu}{\myvecsym{\mu}}

\newcommand{\vc}{\myvec{c}}
\newcommand{\vx}{\myvec{x}}

\newcommand{\gauss}{{\mathcal N}}

\newcommand{\ie}{\textit{i}.\textit{e}., }
\newcommand{\eg}{\textit{e}.\textit{g}. }

\DeclareMathOperator{\relu}{ReLU}
\DeclareMathOperator{\pow}{pow}
\DeclareMathOperator{\CDF}{CDF}

%% file: sections/introduction.tex
\section{Introduction}
\label{secIntroduction}

Probabilistic software analysis methods extend classic static analysis techniques to
consider the effects of probabilistic uncertainty, whether explicitly embedded within
the code -- as in probabilistic programs -- or externalized in a probabilistic input
distribution~\citep{dwyer2017Probabilistic}. Analogously to their classic counterparts,
these analyses aim at inferring the probability of a target event to occur during
execution, \eg reaching a program state or triggering an exception.

For the probabilistic analysis of programs written in a general-purpose programming
language, probabilistic symbolic execution
(PSE)~\citep{geldenhuys2012Probabilistic,filieri2013Reliability,luckow2014Exact}
exploits established symbolic execution engines for the language --
\eg~\citep{cadar2008KLEE,pasareanu2010Symbolic} -- to extract constraints on
probabilistic input or program variables that lead to the occurrence of the target
event. The probability of satisfying any such constraints is then computed via model
counting~\citep{filieri2013Reliability,aydin2018Parameterized} or inferred via solution
space quantification methods~\citep{borges2014Compositional,borges2015Iterative},
depending on the types of the variable and the characteristic of the constraints, and
the probability distributions. Variations of PSE include incomplete analyses inferring
probability bounds from a finite sample of program paths executed
symbolically~\citep{filieri2014Statistical}, methods for non-deterministic
programs~\citep{luckow2014Exact} and data structures~\citep{filieri2015Model}, with
applications to reliability~\citep{filieri2013Reliability},
security~\citep{phan2017Synthesis}, and performance analysis~\citep{chen2016Generating}.
While PSE can solve more general inference problem, the overhead of symbolic execution
is typically justified when the probability of the target event is \emph{very low} (rare
event) or high accuracy standards are required, \eg for the certification purposes of
\emph{safety-critical} systems.

A core element of PSE is to compute the probability of certain variables satisfying a
constraint under the given input probability distribution. In this paper, we focus on
estimating the probability of satisfying numerical constraints over floating-point
variables. For limited classes of constraints and input distributions, analytical
solutions or numerical integration can be computed~\citep{gehr2016PSI,bueler2000Exact}.
However, these methods become inapplicable for more complex classes of constraints or
intractable for high-dimensional problems.

Monte Carlo methods provide a more general and scalable alternative for these estimation
problems. These methods estimate the probability of constraint satisfaction by drawing
samples from the input distribution and estimating the satisfaction probability as the
ratio of samples that satisfy the constraints. Nonetheless, while theoretically
insensitive to the dimensionality of problems, care must be taken to apply direct Monte
Carlo methods in quantifying the probability of \emph{rare events}, \ie when the
probability of satisfying required constraints is extremely small.

To improve the accuracy of the estimation in the presence of low satisfaction
probability, recent work~\citep{borges2014Compositional,borges2015Iterative} uses
interval constraint propagation and branch-and-bound techniques to partition the input
domain of a program into sub-regions that contain \emph{only}, \emph{no}, or \emph{in
part} solutions to a constraint. This step analytically eliminates uncertainty about the
regions containing \emph{only} or \emph{no} solutions, requiring estimation to be
performed only for the remaining regions. The local estimates computed within these
regions are then composed using a stratified sampling scheme: the probability mass from
the input distribution enclosed within each region serves as the weight of the local
estimate, effectively bounding the uncertainty that it propagates through the
composition. However, the performance of this method degrades exponentially with the
dimensionality of the input domain, and it requires an analytical form for the
cumulative distribution function of the input distribution to compute the probability
mass enclosed within each region. Since the cumulative distribution function of most
correlated distributions is not expressible in analytical form, the numerical programs
that can be currently analyzed with PSE are restricted to those with independent inputs.
In this paper, we propose SYMbolic Parallel Adaptive Importance Sampling (SYMPAIS), a
new inference method for the estimation of the satisfaction probability of numerical
constraints that exploits adaptive importance sampling to allow the analysis of programs
processing high-dimensional, correlated inputs. SYMPAIS does not require the
computability of the input cumulative density functions, overcoming the limitations of
current state-of-the-art alternatives relying on stratified sampling. We further
incorporate results from constraint solving and interval constraint propagation to
optimize the accuracy and convergence rate of the inference process, allowing it to
scale to handle higher-dimensional and more general input distributions. We implemented
SYMPAIS in a Python prototype and evaluated its performance on a set of benchmark
problems from different domains.

%% file: sections/background.tex
\section{Background}\label{chap:background}
This section recalls program analysis and
mathematical results required to ground our contribution and details the limitations of the state of the art we aim to tackle.

\vspace{-2mm}
\subsection{Probabilistic Symbolic Execution}\label{sec:background:ppa}

Probabilistic symbolic execution (PSE)~\citep{geldenhuys2012Probabilistic} is a static
analysis technique aiming at quantifying the probability of a target event occurring
during execution. It uses a symbolic execution engine to extract conditions on the
values of inputs or specified random variables that lead to the occurrence of the target
event. It then computes the probability of such constraints being satisfied given a
probability distribution over the inputs or specified random variables. These
constraints are called \emph{path conditions} because they uniquely identify the
execution path induced by an input satisfying them~\citep{king1976Symbolic}.

\begin{minipage}[c]{0.95\columnwidth}
\vspace{5mm}
\begin{lstlisting}[
    language=Java,
    style=mystyle,
    caption={Example code snippet for an example safety monitor of an autopilot navigation system.},
    label=listing:safety]
// Probabilistic profile
altitude ::= Gaussian(8000, 100);
obstacle_x, obstacle_y ::= Gaussian(
    [-2, -2],
    [[0.2, 0.1], [0.1, 0.2]]);
// Program
if (altitude <= 9000) { ...
    if (Math.pow(obstacle_x, 2) +
        Math.pow(obstacle_y, 2) <= 1) {
        callSupervisor();
    ...}
} else { callSupervisor(); }
\end{lstlisting}
\vspace{5mm}
\end{minipage}

Consider the simplified example in~\cref{listing:safety}, adapted
from~\citep{borges2014Compositional} using a Java-like syntax and hypothetical random
distributions for the input variables. The snippet represents part of the safety
controller for a flying vehicle whose purpose is to detect environmental conditions --
excessive altitude or collision distance of an obstacle -- that may compromise the
crew's safety and call for a supervisor's intervention. The purpose of the analysis is
to estimate the probability of invoking \texttt{callSupervisor} at any point in the
code. Safety-critical applications may require this probability to be very small (\eg $<
10^{-7}$) and to be estimated with high accuracy. The symbolic execution of the snippet,
where random variables are marked as symbolic, would return the following two path
conditions~(PCs), corresponding to the possible invocations of \texttt{callSupervisor}:
$PC_0$: $\texttt{altitude} > 9000$; and $PC_1$: $\texttt{altitude} \leq 9000 \land
\pow(\texttt{obstacle\_x}, 2) + \pow(\texttt{obstacle\_y}) \leq 1$.

The probability of satisfying a path condition $PC$ can be computed based on the distributions assigned to the symbolic variables as in~\cref{eqn:event_prob1} (for simplicity, in the remainder of the paper we assume a probability distribution is specified for every symbolic variable or vector of symbolic variables):
\begin{align}
    \label{eqn:p_pc_def}
    &  p_{PC} := \Pr(x \models PC)
    = \int_{\vx} \ind_{PC}(\vx) p(\vx)\, d\vx \\
    &\approx  \frac{1}{N} 
    \sum_{i=1}^N \ind_{PC}(\vx^{(i)}) =: \hat{p}^{DMC},
    \; \text{where} \, \vx^{(i)} \sim p(\vx)\label{eqn:event_prob1}
\end{align}
\noindent where \(\ind_{PC}(\vx)\) denotes the indicator function, which returns 1 if $\vx \models PC$, that is $\vx$ satisfies $PC$, and 0 otherwise. 
For clarity, we will use $\bar{p}(\vx)$ to denote the truncated distribution satisfying the constraints, \ie, $\bar{p}(\vx):=\ind_{PC}(\vx) p(\vx)$.

Because analytical solutions to the integral are in general intractable or infeasible,
Monte Carlo methods are used to approximate $p_{PC}$, as formalized in
Equation~\eqref{eqn:p_pc_def}. When the samples $\vx^{(i)}$ are generated independently
from their distribution $p(\vx)$, Equation~\eqref{eqn:event_prob1} describes a
\emph{direct Monte Carlo (DMC)} integration (also referred to as \emph{hit-or-miss}), 
which is an unbiased estimate of the desired probability and its variance
$\hat{p}^{DMC}(1-\hat{p}^{DMC})/N$ is a measure of the estimator convergence, which can
be used to compute a probabilistic accuracy bound -- \ie the probability of the estimate
deviating from the actual (unknown) probability by more than a positive accuracy $\epsilon>0$~\citep{saw1984Chebyshev}.

Since the path conditions are disjoint~\citep{king1976Symbolic} (\ie $\vx \models PC_i
\land \vx \models PC_j \Rightarrow i=j$), an unbiased estimator for the probability of
the target event to occur through any execution path is $\hat{p}_{PC} = \sum_i
{\hat{p}_{PC_i}}$ over all the $PC_i$ reaching the target event.  

Specialized model counting or solution space quantification methods to solve the
integral in Equation~\eqref{eqn:event_prob1} for PSE application have been proposed for
linear integer constraints~\citep{filieri2013Reliability}, arbitrary numerical
constraints~\citep{borges2014Compositional, borges2015Iterative}, string
constraints~\citep{aydin2018Parameterized}, bounded data
structures~\citep{filieri2015Model}. In this work, we focus on the probabilistic
analysis of program processing numerical random variables.

\subsection{Compositional Solution Space Quantification}\label{secQCoral}

\citet{borges2014Compositional} proposed a compositional Monte Carlo method to estimate
the probability of satisfying a path condition over nonlinear numerical constraints with
arbitrary small estimation variance -- we will refer to this method as \texttt{qCoral}. 
The integrand function in Equation~\eqref{eqn:event_prob1} is an indicator function
returning $1$ for variable assignments satisfying a path condition $PC$, and $0$
otherwise. Such function is typically ill-conditioned for standard quadrature
methods~\citep{quarteroni2007Numerical} and may suffer from the curse of dimensionality
when the number of symbolic variables grows; the ill-conditioning and discontinuities of
the integrand may also lead to high-variance for Monte Carlo estimators, and particular
care should be placed when dealing with low-probability constraints. \texttt{qCoral}
combines insights from program analysis, interval constraint propagation, and stratified
sampling to mitigate the complexity of the integration problem and reduce the variance
of its estimates.

\noindent\textbf{Constraint slicing and compositionality.} As already recalled, the path
conditions of a program are mutually exclusive. Therefore the probability estimates of a
set of path conditions leading to a target event can be added algebraically -- the mean
of the sum being the sum of the means, while the variance of the sum can be bounded from
the variance of the individual summands~\citep{borges2014Compositional}. A second level
of compositionality is achieved in qCoral within individual path conditions via
\emph{constraint slicing}. A path condition is the conjunction of atomic constraints on
the symbolic variables. Two variables depend directly on each other if they appear in an
atomic constraint. The reflexive and transitive closure of this dependency relation
induces a partition of the atomic constraints that groups together all and only the
constraints predicating on (transitively) dependent
variables~\citep{filieri2013Reliability}.

Because each group of independent constraints predicate on a separate subset of the
program variables, its satisfaction probability can be estimated independently from the
other groups. The satisfaction probability of the path conditions is then computed using
the product rule to compose the estimates of each independent
group~\citep{shao2008Mathematical}. Besides enabling independent estimation processes to
run in parallel, constraint slicing can potentially reduce a high-dimensional
integration to the composition of low-dimensional ones -- on independent subsets of the
symbolic variables, in turn leading to shorter estimation time and higher accuracy for a
fixed sampling budget~\citep{borges2014Compositional}.

\noindent\textbf{Interval constraint propagation and stratified sampling.} To further
reduce the variance of the probability estimates of each independent constraint,
\texttt{qCoral} uses interval constraint propagation and branch-and-bound
methods~\citep{granvilliers2006RealPaver} to find a disjoint union of
\emph{n-dimensional boxes} that reliably encloses all the solutions of a constraint --
where $n$ is the number of variables in the constraint. Regions of the input domain
outside the boxes are guaranteed to contain no solutions of the constraint
($\ind_{\cdot}(\cdot)=0$). A box is classified as either an \emph{inner box}, which
contains only solutions, or an \emph{outer box}, which may contain both solutions and
non-solutions. Boxes are formally the conjunction of interval constraints bounding each
of the $n$ variables between a lower and an upper bound: $\bigwedge_{i=0}^n lb_i \leq
x_i \leq ub_i$.

Because the boxes are disjoint, the probabilities of satisfying a constraint $C$ from
values sampled from each box can be composed via \emph{stratified sampling} as the
weighted sum of the local estimates, weighted by the cumulative probability mass
enclosed within the corresponding box~\citep{owen2013Monte}. However, since the inner
boxes contain only solutions, the probability of satisfying $C$ from values sampled from
an inner box is always 1 -- no actual sampling required and, consequently, no estimation
variance to propagate. Sampling and variance propagation is instead required only for
the outer boxes, as per~\cref{eqn:stratified}:
\begin{align}
	\bar{x}_C = \sum_{i=1}^{|O|} p(O_i) \bar{x}_{C \land O_i} + \sum_{j=1}^{|I|} p(I_i), \,
	\bar{v}_C = \sum_{i=1}^{|O|} p(O_i)^2 \bar{v}_{C \land O_i},
	\label{eqn:stratified}
\end{align}
\noindent where $O$ and $I$ are the sets of outer and inner boxes, respectively.
$p(\cdot)$ is the cumulative probability mass in a box, and $\bar{x}_c$ and $\bar{v}_c$
represent the mean and variance of the direct Monte Carlo estimates for constraint $c$.
For independent input variables (as assumed in \texttt{qCoral}), the cumulative
probability mass enclosed in a box is the product of $\CDF(ub_i) - \CDF(lb_i)$ for all
the variables $x_i$ defining the box. Sampling from within a box is possible if the
distribution of a variable $x_i$ can be truncated within the interval $[lb_i, ub_i]$.

\begin{figure}[htb]
\centering
    \includegraphics[width=0.8\columnwidth]{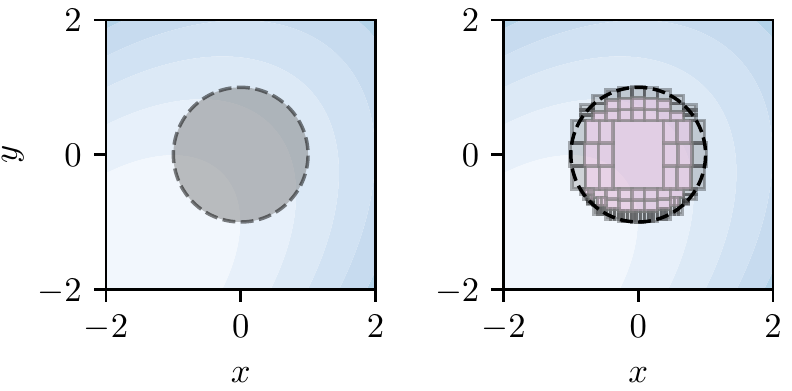}
    \caption[Illustration of the interval boxes produced with the constraint solver
    (RealPaver)]{Left: solution space of $x^2 +y^2 \leq 1$. Right: inner and outer boxes
    produced by RealPaver, in pink and gray, respectively.} \Description{Illustration of
    the interval boxes produced with the constraint solver (RealPaver)}
    \label{fig:cube2d_boxes}
\end{figure}

Stratified sampling with interval constraint propagation can lead to a significant
variance reduction in the aggregated estimate by reducing the uncertainty only to the
regions of the domain enclosed within the outer boxes, potentially avoiding sampling
from large regions of the domain that can be analytically determined as including only
or no solutions. Because boxes can be iteratively refined up to arbitrary accuracy,
there is a trade-off between the target size (and, consequently, weight) of the boxes
and their number (since each outer box requires a Monte Carlo estimation process).

\af{The example can be removed if we need space, which seems likely}
\noindent\emph{Example.} Consider the constraint $x^2 + y^2 \leq 1$ from the example
in~\cref{listing:safety} (variable names abbreviated). Performing interval constraint
propagation with RealPaver~\citep{granvilliers2006RealPaver} -- an interval constraint
solver supporting conjunctive, nonlinear inequality constraints used in \texttt{qCoral}
-- with the initial input domain $x, y \in [-2,2]$, we obtained the outer and inner
boxes depicted in~\cref{fig:cube2d_boxes} in gray and pink, respectively.

In~\cref{fig:cube2d_boxes}, a large region of the domain falls outside the boxes since
it contains no solutions. Hence, the probability of satisfying the constraint for values
in this region is $0$. Similarly, the probability of satisfying the constraint with
inputs from an inner box is $1$. Therefore, uncertainty is bounded within the outer
boxes, and estimation proceeds sampling from their truncated distributions and
aggregating the result via stratified sampling.

\subsection{Limitations of qCoral}\label{secQCoralLimitations}
\texttt{qCoral} can produce scalable and accurate estimates for the satisfaction probability for constraints
that %
\begin{inparaenum}[1)]
    \item have low dimensionality or can be reduced to low-dimensional subproblems via
    constraints slicing,
    \item are amenable to scalable and effective interval constraint propagation, and
    \item whose input distribution have CDFs in analytical form and allows efficient
    sampling from their truncated distributions.
\end{inparaenum}
These constraints typically do not hold for \emph{high-dimensional and correlated input distributions}.

\noindent\textbf{Constraint slicing} assumes that all the inputs are probabilistically independent, with dependencies among variables arising only from computational operations (\eg $\texttt{if}\;(x + y >0){\dots}$ ).
Support for correlated variables requires changing the dependency relation to include also all correlated variable pairs. This may reduce the effectiveness of constraint slicing in reducing the dimensionality of the integration problems.

\noindent\textbf{Interval constraint propagation} contributes to reducing estimation variance by pruning out large portions of the input domain that do not contain solutions of a constraint and producing small-size outer boxes to bound the variance propagated from in-box local estimates.
However, the complexity of this procedure grows exponentially with the dimensionality of the problem, rendering it ineffective when, after constraint slicing, the number of variables appearing in an independent constraint is still large, \eg due to correlated inputs that cannot be separate.
The effectiveness of interval constraint propagation for nonlinear, non-convex constraints also varies significantly for different formulations of the constraint (\eg $x^2$ vs. $x \times x$) and may require manual tuning for optimal results~\citep{granvilliers2006RealPaver}.

\noindent\textbf{Stratified sampling} requires analytical solutions of the input CDFs, as well as the ability to sample from truncated distributions. Both requirements are generally unsatisfiable for correlated input variables, whose CDF cannot be computed in closed form.
The lack of an analytical CDF would require a separate Monte Carlo estimation problem to quantify the probability mass enclosed within each box and an analysis of how the corresponding uncertainty propagates through the stratified sampling and the composition operators of \texttt{qCoral}.
Additionally, sampling from a truncated distribution typically relies on the computation of both the CDF and the inverse CDF of the original distribution, which is inefficient without an analytical form of these functions.

In summary, the main variance reduction strategies of \texttt{qCoral} based on interval constraint propagation and stratified sampling are not applicable for all but trivial correlated input distributions. 
Constraint slicing can be extended with probabilistic dependencies among input variables, but this results in smaller dimensionality reduction, with exponential impact on interval constraint propagation even when the CDFs of correlated inputs can be computed analytically.

\subsection{Importance Sampling}\label{secBgdImportanceSampling}

The indicator function in Equation~\eqref{eqn:event_prob1} return 1 only within the
regions of the input domain satisfying a constraint (\eg only within the circle in
Figure~\ref{fig:cube2d_boxes}). When this region encloses only a small probability mass,
direct Monte Carlo methods sampling from the input distribution $p(\vx)$ may struggle to
generate enough samples that satisfy the constraint, and therefore fail to estimate the
quantity of interest $p_{PC}$. We discussed before how \texttt{qCoral} uses interval
constraint propagation and stratified sampling to prune out regions of the domain
that contain no solutions, sampling within narrower boxes containing a larger portion of solutions.

An alternative method to improve statistical inference in this problem is
\emph{importance sampling} (IS). Instead of sampling from the input distribution
$p(\vx)$, IS generates samples from a different \emph{proposal} distribution -- $q(\vx)$
-- that overweighs the important regions of the domain, \ie the regions containing
solutions in our case. Because the samples are generated from a different distribution
than $p(\vx)$, the computed statistics need to be re-normalized as
in~\cref{eqn:is_def}:
\begin{align}
	\label{eqn:is_def}
    &  p_{PC} :=  \int_{\vx} \ind_{PC}(\vx) p(\vx)\, d\vx = \int_{\vx} \frac{\ind_{PC}(\vx) p(\vx)}{q(\vx)} q(\vx)\, d\vx \\
    &\approx  \frac{1}{N} 
    \sum_{i=1}^N \frac{\ind_{PC}(\vx^{(i)}) p(\vx^{(i)})}{q(\vx^{(i)})}  =: \hat{p}^{IS},  
    \; \text{where} \, \vx^{(i)} \sim q(\vx).
\end{align}
While any distribution $q(\vx) > 0$ over the entire domain guarantees the estimate will
eventually converge to the correct value, an optimal choice of $q(\vx)$ determines the
convergence rate of the process and its practical efficiency. In our context of
estimating the probability of satisfying path conditions \(PC\), the \emph{optimal proposal
distribution} \(q^*(\vx)\) is \emph{exactly} the truncated, normalized distribution
\(p(\vx)\) satisfying \(PC\),
\[
  q^*(\vx) = \frac{1}{p_{PC}}p(\vx)\ind_{PC}(\vx).\label{eqn:optimal-proposal}
\]
In general, it is infeasible to sample from \(q^*(\vx)\) as it requires the
calculation of $p_{PC}$ which is exactly our target. Fortunately, as we will demonstrate in~\cref{secSympaisIS}, a proposal distribution found via adaptive refinement can allow us to achieve near-optimal performance.

\emph{In this paper, we propose a new inference method to estimate the satisfaction probability of numerical constraints on high-dimensional, correlated input distributions.}
Our method does not require analytical CDFs and can replace \texttt{qCoral}'s variance
reduction strategies to analyze constraints where these are not applicable. 
Our method combines results from constraint solving and adaptive estimation to produce near-optimal
proposal distributions aiming at computing high-accuracy estimates suitable for the analysis of low-probability constraints.

%% file: sections/sympais.tex
\section{SYMPAIS: SYMbolic Parallel Adaptive Importance Sampling}
\label{sec:SYMPAIS}

In this section, we introduce our new solution space quantification method for
probabilistic program analysis: SYMbolic Parallel Adaptive Importance
Sampling~(SYMPAIS).

\begin{figure}[htb]
\centering
    \includegraphics[width=1.0\columnwidth]{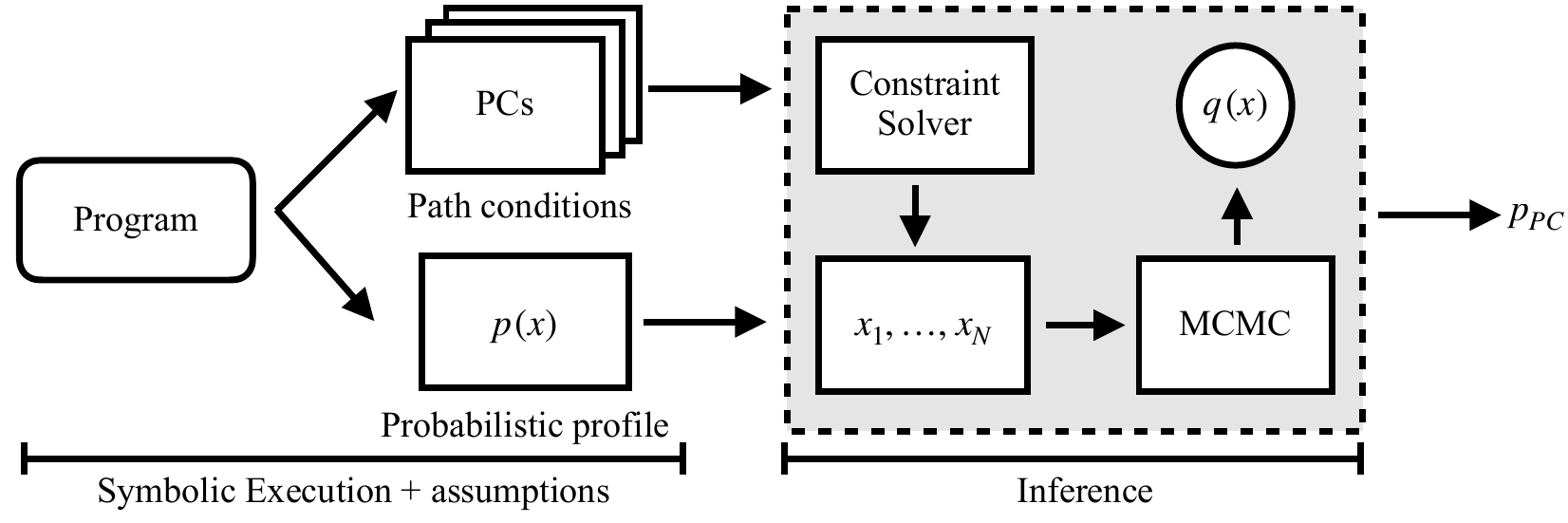}
    \caption[Overview of SYMPAIS]{Overview of SYMPAIS.}
    \Description[Overview of SYMPAIS]{Overview of SYMPAIS.}
    \label{fig:sympais}
\end{figure}

\Cref{fig:sympais} gives an overview of SYMPAIS's workflow. Following the
probabilistic symbolic execution approach, the path conditions leading to the occurrence
of a target event are extracted using a symbolic execution engine. For simplicity, we
assume that each (vector of) symbolic variables is associated with a probability
distribution, either provided explicitly by the user or extracted from the code, where
it has been specified via convenient random generators. Independent variables are
associated with univariate probability distributions. Vectors of correlated variables
are associated with multivariate probability distributions. For example, in
\cref{listing:safety}, \texttt{altitude} is associated with a univariate Gaussian
distribution with location $8000$ and scale $500$, while $<\texttt{obstacle\_x},
\texttt{obstacle\_y}>$ are distributed as a bivariate Gaussian with location $[-2, -2]$
and covariance matrix $[[0.1, 0.1], [0.1, 0.2]]$. The path conditions are assumed to
have been sliced as in~\citep{filieri2013Reliability,borges2014Compositional}, where the
dependency relation is augmented with pairwise dependency between the correlated
variables, besides the dependencies induced by the program control and data flows. The
probability of satisfying each independent constraint is quantified in the inference
phase, which is the focus of this work.

\begin{algorithm}[thb]
\caption{Symbolic Parallel Interacting Markov Adaptive Sampling (SYMPAIS)}
\label{algo:sympais}
\begin{algorithmic}[1]
    \State Given $C, p(\vx)$  \Comment{SymEx(P), domain knowledge}\label{alg:pimais-prog:extract}
    \State $\overline{p}(\vx) \gets p(\vx)\ind_{C}(\vx)$\label{alg:pimais-prog:target}
    \State $\vx_{\texttt{init}} \gets models(C)$\Comment{Constraint solver}\label{alg:pimais-prog:initial}
    \State Initialize the proposal distribution \(q_{n,0}\) for \(n = 1,\dots,N\).
    \For{$t \gets 1,\dots,T$}\label{alg:pimais-prog:loop-start}\Comment{Run PI-MAIS for $T$ iterations}
        \State Update $N$ proposal distributions \(q_{1:N,t}\) 
        using MCMC\label{alg:pimais-prog:adapt}
        \State Draw \(M\) samples from each proposal distribution,  
        \begin{align}
            \vx^{(m)}_{n, t} \sim q_{n,t}(\vx), 
            \quad \text{for} \; m = 1:M, \text{and} \, n = 1:N.
        \end{align}
        \State Compute importance-sampling weights, 
        \begin{align}
            w_{n,t}^{(m)} \gets 
            \frac{\overline{p}(\vx_{n,t}^{(m)})}
                 {\frac{1}{N}
                  \sum_{j=1}^N {q_{j,t}(\vx_{n,t}^{(m)})}}
        \end{align}
    \EndFor\label{alg:pimais-prog:loop-end}
    \State Obtain PI-MAIS esitmate, 
    \begin{align}
        \hat{p}_{\text{PIMAIS}} &\gets 
    \frac{1}{T \cdot N \cdot M}\sum_{t=1}^T\sum_{n=1}^N\sum_{m=1}^M w^{(m)}_{n, t}, \label{alg:pimais-prog:final-estimate}\\ 
    \hat{v}^2_{\text{PIMAIS}} &\gets \frac{1}{T \cdot N \cdot M}\sum_{t=1}^T\sum_{n=1}^N\sum_{m=1}^M (w^{2(m)}_{n, t} - \hat{p}_{\text{PIMAIS}})^2 \label{alg:pimais-prog:final-variance}
    \end{align}
    \State \Return $\hat{p}_{\text{PIMAIS}}$, $\hat{v}^2_{\text{PIMAIS}}$
\end{algorithmic}
\end{algorithm}

The main steps of SYMPAIS are summarized in Algorithm~\ref{algo:sympais}. SYMPAIS takes
as input a constraint $C$, which may be a path condition of $P$ or an independent
portion of the path condition after constraint slicing, $C$ is assumed to be the
conjunction of inequalities (or equalities) on numerical functions of the inputs. In
addition to the constraint $C$, SYMPAIS requires specifying a probability distribution
$p(\vx)$ over the symbolic variables in the program. Such distribution can be provided
by the user or specified in the code via convenient random generators. For simplicity,
we refer to the probability distribution over all of the symbolic variables as the input
distribution.

\noindent\textbf{Overview.} The core part of SYMPAIS is the adaptive importance sampling process implemented in the for loop at~\cref{alg:pimais-prog:loop-start}. The goal of this process is to iteratively refine an importance sampling proposal that maximizes sample efficiency, \ie it is very likely to generate sample points within the solution space of the input constraint $C$. Due to the wide range of possible constraint forms (\eg linear, non-linear, non-convex) and of different types of distributions, optimal proposal distributions cannot be obtained analytically. It is instead approximated via a hierarchical probability distribution whose parameters are iteratively refined via a Markov Chain Monte Carlo (MCMC) algorithm~\citep{metropolis1953Equation,neal2011MCMC} to best approximate the intractable optimal proposal. MCMC algorithms generate sequences of samples that, when the process converges to its steady state, the samples are distributed according to a target distribution whose analytical form may be unknown or from which it is not possible or intractably complex to sample directly. The MCMC samples can thus be used to iteratively estimate the parameters of the proposal distribution towards approximating the optimal distribution. The algorithm returns the estimate of the satisfaction probability, as well as the estimator variance. The latter may be used to reason about the dispersion of the estimate, e.g., constructing confidence intervals to decide if more sampling is desirable. Notice however that, similarly to \texttt{qCoral}~\cite{borges2014Compositional}, the estimator variance is centered around the estimate (Formula~\eqref{alg:pimais-prog:final-variance} in Algorithm~\ref{algo:sympais}). This requires enough samples to have been collected for the estimate to stabilize first in order for the variance to represent the estimator dispersion around it.

In the remaining of the section (\cref{secSympaisIS}), we will formally define the adaptive importance sampling strategy of SYMPAIS and the MCMC methods it adopts for the adaptive refinement of the proposal distribution. 
Results from constraint solving will be brought in to mitigate the complexity of the estimation process and accelerate its convergence.
A set of optimizations to improve the practical performance of the methods will be discussed in Section~\ref{sec:optimization},
while implementation details and an experimental evaluation will be reported in Section~\ref{secEvaluation}.

\noindent\textbf{Running example.}
To illustrate the different features of SYMPAIS, we will use a 3-dimensional, nonlinear
and non-convex constraint -- \texttt{torus} --, which is defined in
Equation~\eqref{eqn:torus}:
\begin{equation}
    (\sqrt{x^2 + y^2} - R)^2 + z^2 \le r^2,\label{eqn:torus}
\end{equation}
\noindent with the constant parameters \(R = 3\), \(r = 1\). At first, we will associate
to each of the three variables an independent univariate Gaussian distribution: \ie $x,
y, z \sim \gauss(0, 0.5)$. We will later generalize the method to correlated inputs.

\subsection{SYMPAIS Adaptive Importance Sampling}\label{secSympaisIS}

As recalled in Section~\ref{secBgdImportanceSampling},
importance sampling (IS) methods aim at constructing a proposal distribution $q(\vx)$
that increases the likelihood of generating samples that satisfy a constraint $C$.
This allows us to focus the estimation problem to the regions of the input domain that satisfy $C$, while avoiding the need to find a stratification of the input domain and computing the inverse CDFs for the distribution truncation that prevent the use of \texttt{qCoral} for high-dimensional and correlated input distributions.
\begin{figure}[ht]
    \centering
    \includegraphics{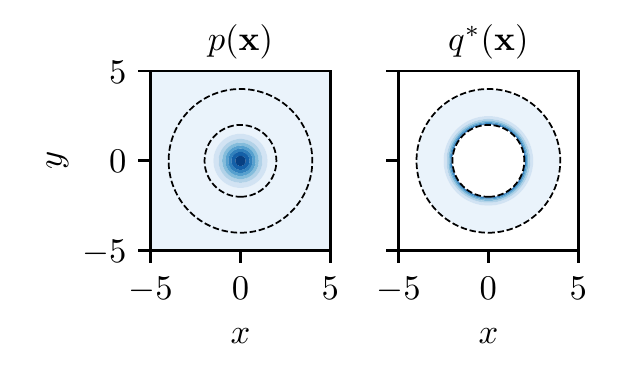}
    \caption{Left: input distribution $p(\vx)$ for torus projected on the x-y plane. 
    Right: the optimal importance sampling proposal $q^*(\vx)$. The intensity of the blue shadowing is proportional to the probability density. The solution space lays between the dashed circles.}
    \Description{Visualizing the input distribution and optimal proposal}
    \label{fig:torus-independent-optimal-proposal}
\end{figure}

The choice of a proposal distribution \(q(\vx)\) largely affects the efficiency of IS. 
However, it is usually difficult to obtain an analytical form for the theoretically optimal $q^*(\vx)$ or to sample from.  
Instead, we use an adaptive scheme to iteratively refine a proposal distribution to approximate $q^*(\vx)$.

\subsubsection{Adaptive proposal refinement}

To construct and refine the IS proposal distribution, SYMPAIS adapts the parallel interacting Markov adaptive sampling (PI-MAIS) schema defined in~\citep{martino2017Layered}.
The proposal distribution in PI-MAIS is a hierarchical model parameterized by $N$ \emph{sub-proposals} \(q_1, \dots, q_N\).
To sample from the proposal distribution, we first choose a sub proposal \(q_i\) and then draw samples \(\vx\) 
from \(q_i\). Together, the sub-proposals form a mixture distribution. PI-MAIS adapts the sub-proposals to the target distribution by running parallel-chain MCMC (Line~\ref{alg:pimais-prog:adapt}) so that it can form efficient proposal distributions for target distributions that are multimodal and non-linear\footnote{We provide an executable notebook with more details on PI-MAIS in the open-source code~\citep{luo2021github}.}.

In this paper, we use sub-proposals parameterized by Gaussian distributions $q_i(\vx) = \gauss(\vx; \vmu_i, \Sigma)$.
The probability density function (PDF) for the proposal distribution is then given by a Gaussian: %
\begin{equation*}
    q(\cdot) = \frac{1}{N} \sum_{i=1}^{N} \gauss(\cdot, \vmu_i, \Sigma),
\end{equation*}
\noindent where the mean vectors $\left\{\vmu_i\right\}_{i=1}^N$ are adapted by running $N$ parallel MCMC samplers so that the proposal distribution approximates more accurately $q^*(\vx)$. At each step $t$, a sampler produces a set of samples $\{\vx_{n,t}\}$. The proposal distribution at step $t$ is:
\begin{equation*}
    q_t(\cdot) = \frac{1}{N} \sum_{n=1}^{N} q_{n,t}(\cdot) =
    \frac{1}{N} \sum_{n=1}^{N} \gauss(\cdot | \vx_{n,t}, \Sigma).
\end{equation*}

When the refinement process stabilizes, the estimate for the probability of satisfying the constraint $C$ given the input distribution $p(\vx)$ (\ie the solution of the integration problem in Equation~\eqref{eqn:p_pc_def}) is:
\begin{equation*}
    \hat{p}^{PIMAIS} \approx \frac{1}{T}\frac{1}{N}\frac{1}{M}\sum_{t=1}^{T}\sum_{n=1}^{N}\sum_{m=1}^{M} w_{t,n,m},\,  w_{t,n, m} = \frac{\overline{p}(\vx^{(m)}_{n,t})}{q_{t}(\vx^{(m)}_{n,t})},
\end{equation*}
where $\vx^{(m)}_{n,t}$ are samples drawn from $q_{n,t}(\vx)$. Please refer to~\citep{martino2017Layered} for the convergence proofs of the PI-MAIS scheme.

To update the proposal distribution, we implemented two MCMC samplers in SYMPAIS:  random-walk Metropolis-Hastings (RWMH) and Hamiltonian Monte Carlo (HMC). The former provides a general procedure that only requires the ability to evaluate the density of the constrained input distribution $\bar{p}(\vx)=p(\vx)\ind_{C}$ for a given value $\vx$. 
The latter requires the $p(x)$ to be differentiable and exploits the gradient information to achieve higher efficiency in many cases, especially for higher dimensional problems. For space reason, in the remainder of this section we will mostly focus on RWMH while additional details on our HMC implementation are provided in~\citep{luo2021sympais}.

\noindent\textbf{Random-Walk Metropolis-Hasting for adaptation.} 

MCMC methods generate a sequence of samples
where each sample is via a probabilistic transition from its predecessor.
Random-Walk Metropolis-Hasting (RWMH)~\citep{owen2013Monte,metropolis1953Equation} is an MCMC algorithm
where the next sample $\vx'$ is generated from its predecessor $\vx$ from a
proposal distribution (or proposal kernel) $\kappa(\vx' | \vx)$.
The newly proposed sample $\vx'$ is accepted and added to the sequence randomly with probability  $\alpha = \min \left(1, \frac{\overline{p}(\vx')\kappa(\vx \mid \vx')}{\overline{p}(\vx)\kappa(\vx' \mid \vx)}\right)$, 
otherwise the new sample is rejected and $\vx$ is retained.

For RWMH within SYMPAIS, we use $\kappa(\vx' |\vx) = \gauss(\vx'; \vx,  \Sigma)$,
\ie the next candidate sample is generated by adding a white Gaussian noise with covariance $\Sigma$ to the current sample $\vx$.
After the generation process converges at steady state, a value $\vx$ should appear in the sequence with a frequency proportional to its probability in the target distribution $\overline{p}(\vx)$. Because $\overline{p}(\vx)$ is zero outside the solution space of the constraint $C$, all the samples that do not satisfy $C$ will be rejected. High rejection rate slows down the convergence and can be mitigated by tuning $\Sigma$, using a different proposal kernel~\citep{owen2013Monte}, or switching to more sophisticated methods to generate the next sample, such as a Hamiltonian proposal.

\subsubsection{SYMPAIS estimation process}

Each of the parallel MCMC processes used to refine the importance sampling proposal requires an initial value $\vx_{init}$ to start from. In theory, any point from the input domain can be chosen to start the MCMC processes. However, principled choices of $\vx_{init}$ can speed up the converge of the Markov chain to a steady state and reduce the sample rejection rate. The choice of the initial points is particularly important when the constrained distribution $\overline{p}(\vx)$ is multimodal -- \ie its density function has two or more peaks --, either because the original input distribution $p(\vx)$ is itself multimodal or because the restriction to the solution space of $C$ induces multiple modes in $\overline{p}(\vx)$. We will discuss the problem of multiple modes and SYMPAIS's mitigation strategies in Section~\ref{sec:optimization} while focus here on SYMPAIS's use of constraint solving to initialize the MCMC processes.

In statistical inference literature, the initial sample of an MCMC process is typically randomly assigned by a value within the input domain. 
However, if the satisfaction probability of the constraint $C$ is small, randomly generating a value of $\vx$ that satisfies the constraint may require a large number of attempts. 
Instead, we use a \emph{constraint solver} (Z3~\citep{demoura2008Z3} in this work) to generate one or more models for the constraint $C$ to seed the MCMC processes.

\begin{figure}[htb]
\centering
    \includegraphics{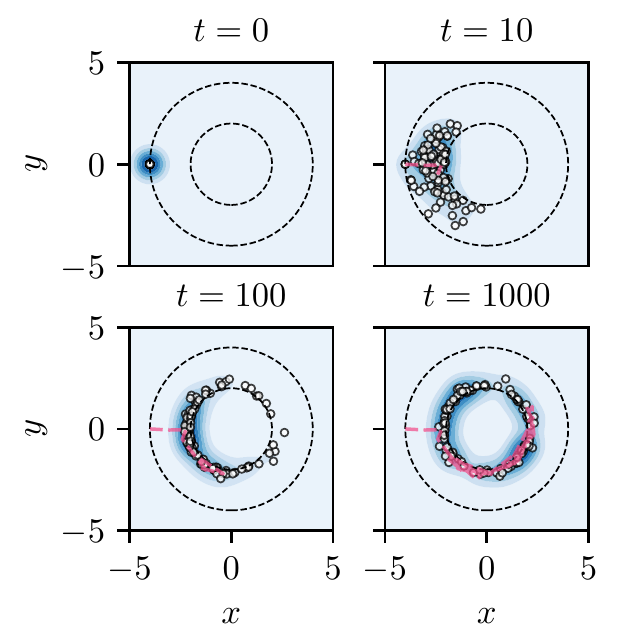}. %
    \caption[Graphical illustration of learning the adaptive proposal in SYMPAIS.]{Graphical illustration of learning the adaptive proposal in SYMPAIS.}
    \Description[Graphical illustration of learning the adaptive proposal in SYMPAIS.]{Graphical illustration of learning the adaptive proposal in SYMPAIS.}
    \label{fig:pimais_demo}
\end{figure}

\Cref{fig:pimais_demo} demonstrates visually the evolution of the proposal distribution through the iterations of the SYMPAIS loop (Line~\ref{alg:pimais-prog:loop-start}) towards the optimal proposal distribution depicted in Figure~\ref{fig:torus-independent-optimal-proposal}. We use a projection of the torus constraint on the x-y plane again as an example. At the beginning ($t=0$), the process is initialized with a solution produced by Z3. The importance sampling proposal distribution is concentrated around that point, where darker shadows of blue represent higher probability density. At iteration $t=10$, the proposal distribution translated towards the inner border of the solution space, where the constrained input distribution $\overline{p}(\vx)$ has a higher density.
The white dots represent the samples $\vx_{n,t}$ from the MCMC processes that are also used to refine the mean vectors of the importance sampling proposal $q_{n,t}$. Proceeding through the iterative refinement, at iteration $t=1000$ the proposal distribution approximates the optimal proposal very closely. The red line in the rightmost subfigure shows a trajectory -- a sequence of values -- generated by one of the MCMC processes, which touches portions of the solution space approximately proportionally to their density in the optimal distribution. 

The accurate approximation of the optimal proposal distribution allows SYMPAIS to effectively sweep the solution space of $C$ and estimate its satisfaction probability.

\noindent\textbf{Correlated input distributions.} The adaptive importance sampling strategy, as well as the MCMC processes described in this section, do not require the input distributions to be independent. Correlated distributions (such as the bivariate Gaussian in~\cref{listing:safety}) can be seamlessly processed by SYMPAIS. The requirement for RWMC is the ability to evaluate the PDF of the distribution, while HMC requires its differentiability. Computing the CDF (and its inverse) as required for \texttt{qCoral}'s stratified sampling does instead involve an integration problem that usually has no analytical solution and requires a separate Monte Carlo integration. SYMPAIS thus complements \texttt{qCoral} to allow the probabilistic analysis of a broader range of programs. We will demonstrate applications of SYMPAIS to correlated input distributions in~\cref{secEvaluation}.

\vspace{-2mm}
\subsection{Optimizations}\label{sec:optimization}
The target distribution of SYMPAIS is $\overline{p}(\vx) = p(\vx)\ind_{PC}(\vx)$, where the indicator function zeroes the input distribution's density outside the solution space of $C$. However, the MCMC processes are not aware of the geometry or location of the solution space of $C$. The volume and shape of the solution space may affect the rejection rate of the processes -- how often the random walk reaches non-solution points -- and may induce multiple modes in $\overline{p}(\vx)$ even if $p(\vx)$ is unimodal. Intuitively, each mode is a peak in the density function of $\overline{p}(\vx)$, which behaves as an attractor for the MCMC processes, requiring a longer time to converge to covering all the modes.

\begin{figure}[ht]
\centering
    \includegraphics{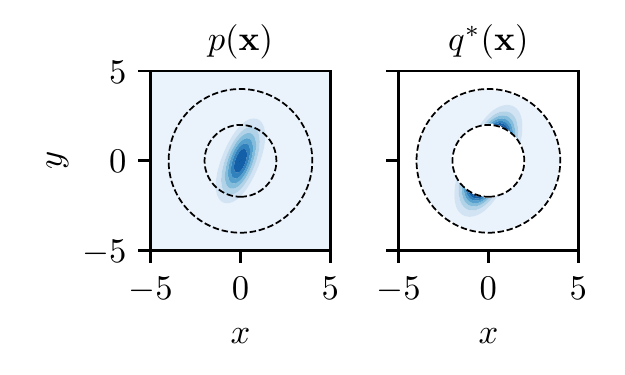}
    \caption[Illustration of the input distribution and optimal proposal distribution]{Left: unimodal correlated input distribution $p(\vx)$. Right: optimal bi-modal proposal distribution $q^*(\vx)$.}
    \Description{Illustration of the input distribution and optimal proposal distribution.}
    \label{fig:torus-correlated-optimal-proposal}
\end{figure}

\noindent\emph{Example.} Figure~\ref{fig:torus-correlated-optimal-proposal} shows how the optimal proposal distribution $q^*(\vx)$ for a unimodal, correlated input distribution $p(\vx)$ -- we used a Student's T distribution with 2 degrees of freedom for the plot -- degenerates into a bimodal optimal proposal when constrained within the solution space of $C$ (torus constraint projected to the x-y plane). An MCMC process initialized in the neighborhood of one of the mode may take a long time before ``jumping'' in the neighborhood of the other mode, having to traverse a low probability path across the non-convex solution space.

While a complete characterization of $C$'s geometry is intractable, in this section, we propose three heuristic optimizations that may mitigate the impact of ill geometries of the solution space on SYMPAIS adaptive importance sampling.

\subsubsection{Diverse initial solutions.}
For an effective importance sampling, the adaptive proposal distribution should capture all the modes of $q^*(\vx)$. Running multiple MCMC processes in parallel increases the chances of at least any of them covering each mode. In the statistical inference literature, each chain is typically initialized with an independent random sample from the input distribution to maximize the chances of reaching all the modes on the whole. However, as discussed before, if the satisfaction probability of $C$ is small, it is unlikely to randomly generate valid solutions and even less likely to also cover multiple modes. 

\begin{figure}[ht]
\centering
    \includegraphics[width=\columnwidth]{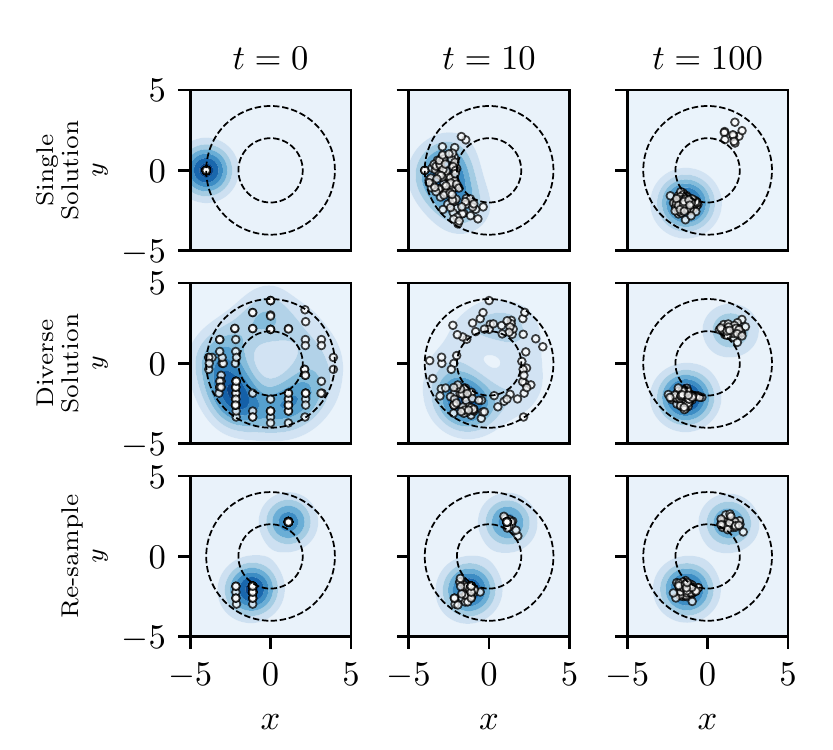}
    \caption{Convergence of the adaptive proposal distribution with different initialization strategies.}
    \Description{Convergence of the adaptive proposal distribution with different initialization strategies.}
    \label{fig:pimais_init}
\end{figure}

Initializing all the chains with a feasible solution generated by the constraint solver may result in the MCMC processes exploring, in a finite time, only the mode closest to the initial solution. We observed this phenomenon in particular for RWMH, but multimodal distributions require longer convergence times also with HMC.  The top row in Figure~\ref{fig:pimais_init} shows the evolution of the adaptive importance sampling proposal $q(\vx)$ of SYMPAIS initialized with a single solution from the constraint solver. This time, the optimal proposal distribution $q^*(\vx)$ is the one on the right-hand side of Figure~\ref{fig:torus-correlated-optimal-proposal}. After $t=100$ iterations, $q(\vx)$ still fails to converge to $q^*(\vx)$, with most of the samples still generated around one of the two modalities -- which have instead the same density in $q^*(\vx)$.

To mitigate this problem, SYMPAIS tries to generate multiple diverse solutions of $C$ to increase the probability of obtaining at least one in the neighborhood of each mode. We explored two different approaches for this purpose. In principle, initial solutions diversity can be achieved using an optimizing solver where each solution is chosen to maximize the distance from all the previous ones. This method is general and flexible (\eg it allows customizing the distance function based on domain-specific information), but computationally heavy for non-linear, non-convex constraints. 

An alternative, more scalable method relies on interval constraint propagation and branch and bound algorithms to single out regions of the input domain that satisfy $C$. In our implementation, we use RealPaver's depth-first search mode with a coarse accuracy~\citep{granvilliers2006RealPaver} ($10^{-2}$ in this example, but the configuration can be tuned based on the length of the domain dimensions) to generate several boxes that contain solutions of $C$. This mode differs from the standard paving used in \texttt{qCoral} because it does not require the computed boxes enclosing all the solutions, but potentially only a subset of them, making it more scalable even for higher-dimensional problems. For each box, SYMPAIS calculates the center and, if it satisfies $C$, adds it to the MCMC initialization points.

An example of the SYMPAIS adaptive proposal distribution using this heuristic is shown in the middle row of Figure~\ref{fig:pimais_init}. The initial solutions cover, with different concentrations, several parts of the solution space of $C$. While the initial iterations result in a very spread proposal $q(\vx)$ -- contrary to the highly concentrated optimal proposal -- the diversity of the initial points allows the adaptive refinement to converge to covering both optimal proposal's modes.

While the heuristics of using depth-first interval constraint propagation may fail to produce solutions at the center of its boxes, or to refine the boxes enough for it to happen, when applicable, it is an efficient alternative to solving heavier optimization problems.

\subsubsection{Re-sampling}
The diversification of the initial solutions described in the previous section uses only information about the constraint to generate diverse initial points. However, the constraint solver is not aware of the underlying input distribution and may generate many solutions that are far from the modes of $q^*(\vx)$. When the $N$ parallel MCMC processes are initialized with solutions taken uniformly at random from those generated by constraint solver, many of these solutions are likely to be far from the modes, therefore requiring longer warmup of the Markov chains to move towards the modes. This can be observed in the wide spread of the proposal $q(\vx)$ for $t=0,10$ in the middle row of Figure~\ref{fig:pimais_init}.

To reduce the number of samples used for warmup, we sample the initial solutions proportionally to their likelihood in the input distribution. Let $\{\vx_i\}_{i=1}^F$ be the initial solutions found by the constraint solver. We sample $N$ initial points $\{\vx'_i\}_{i=1}^N$ from 
\[
    q(\vx') = \sum_{i=1}^F w_i \, \delta_{\vx_i}(\vx'), \quad w_i = p(\vx_i) / \sum_{j=1}^{F} p(\vx_j),
\] to be the initial states for the $N$ parallel MCMC chains.

The initial solutions seeding the MCMC chains now reflect both the location of the solutions space -- from the constraint solver -- and the distribution of the input probability across the solution space -- from the resampling. An example of the effects on the convergence speed of SYMPAIS adaptation is shown in the bottom row of Figure~\ref{fig:pimais_init}.

\subsubsection{Truncated kernel for RWMH}
The proposal kernel for RWMH defined in the previous section generates the next sample by adding Gaussian noise to the current one: $\kappa(\vx' | \vx) = \gauss(\vx'; \vx, \Sigma)$. While arbitrarily concentrated around $\vx$ by the value of $\Sigma$, this kernel may propose new samples far away from the solution space of $C$, which would then be rejected. 

A way to reduce the rejection rate is to replace the kernel with a Gaussian noise truncated within a smaller region of the input domain that contains the solution space of $C$. We obtain this region in the form of the smallest n-dimensional box that contains $C$. Such a box can be efficiently computed when an interval contractor function is defined for $C$~\citep{araya2012Contractor}. Efficient interval contractors are implemented for a broad class of numerical constraints, \eg~\citep{goldsztejnibexlib}, and can be used to compute the smallest box $\mathcal{B}$ that encloses the solutions space of $C$.
The Gaussian proposal kernel of RWMH can then also be replaced by the truncated Gaussian proposal kernel $\kappa_\mathcal{B}(\vx' | \vx) = \gauss_\mathcal{B}(\vx'; \vx, \Sigma)$ to increase the probability of generate candidate samples $\vx'$ that are still solutions of $C$. 

Notice that this optimization does not require us to truncate the input distribution $p(\vx)$, as required by \texttt{qCoral}. Instead, it truncates the uncorrelated Gaussian distribution of the proposal kernel of RWMH, which can be done efficiently. Notably, the rejection rate does not go to zero because $\mathcal{B}$ may be a coarse bounding of the solution space of $C$, which may also be non-convex. Nonetheless, it usually increases the probability of sampling solutions of $C$.

%% file: sections/evaluation.tex
\section{Evaluation}\label{secEvaluation}

In this section, we report an experimental evaluation of SYMPAIS. We include direct
Monte Carlo (DMC) estimation as a baseline, \texttt{qCoral} for the uncorrelated input
distributions, and SYMPAIS and SYMPAIS-H where we configure SYMPAIS to use the RWMH and
the HMC algorithms for the MCMC samples, respectively. We include two geometrical
microbenchmarks to expose the features of SYMPAIS and six benchmarks from path
conditions extracted from a ReLU neural network and subjects from \texttt{qCoral}.
Because the dependability of Monte Carlo estimators' variance depends on the convergence
of their estimates (cf. \cref{sec:SYMPAIS}, overview), we use the relative absolute
error (RAE) to compare the estimates against the ground truth, instead of performing
statistical tests on the variance that are particularly challenging with the rare events
considered in this paper. Details about the experimental environment and set-up are available in the extended version~\cite{luo2021sympais}. Code is available at~\cite{luo2021github}.

\subsection{Geometrical Microbenchmarks}
\subsubsection{Sphere}
The first constraint we consider the d-dimensional sphere: \(C := \left\{\left\lVert\vx
- \vc\right\rVert^2 \le 1 \right\}\), where \(\vx \in [-10, 10]^d \cap \mathbb{R}^d \)
is the input domain, and $\vc \in \mathbb{R}^d$ is the center of the sphere. We use
\(p(\vx) = \gauss(0, I)\) -- \ie uncorrelated Gaussian -- as the input distribution and
set \(\vc = \mathbf{1}\). Despite its simplicity, this problem illustrates the
challenges faced by direct Monte Carlo methods as well as \texttt{qCoral} in
high-dimensionality problems where $C$'s satisfaction probability is small. 

Specifically, as \(d\) increases, the probability of the event happening decreases,
which makes estimation by DMC increasingly challenging. Moreover, the increase in \(d\)
also leads to coarser paving of $d$-dimensional boxes, which reduces the effectiveness of
variance reduction via stratified sampling.

\begin{figure}[t]
    \centering
    \includegraphics[width=\columnwidth]{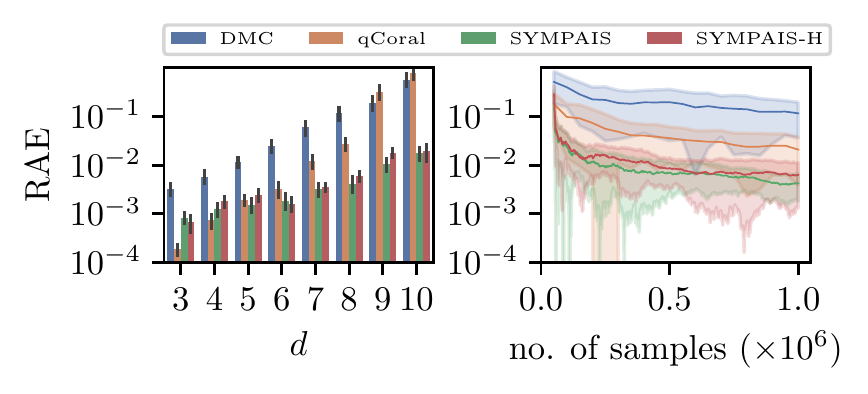}
    \caption{RAE and convergence rates for spheres of different dimensionality.}
    \Description{RAE and convergence rates for spheres of different dimensionality.}
    \label{fig:sphere_result}
\end{figure}

The RAE results are illustrated in~\cref{fig:sphere_result} (left). As expected: DMC
achieves the worst performance throughout all tests. For low-di\-mensional problems ($d
\leq 4$), \texttt{qCoral} is the most efficient, while its performance deteriorates
significantly when the $d$ increases and RealPaver fails to prune out large portions of
the domain that contain no solutions. SYMPAIS's performance is comparable to
\texttt{qCoral} in low dimensions, but up to one order of magnitude more accurate when
the dimensionality grows ($d \geq 8$). 

Figure~\ref{fig:sphere_result} (right) shows the convergence rate of RAE for different
methods over sample size for $d=8$.  SYMPAIS achieves the final RAE of DMC with $<5\%$
of the sampling budget and the final RAE of \texttt{qCoral} with $<10\%$. SIMPAIS-H only
marginally outperforms SYMPAIS for $8 \leq d \leq 10$. The improvement in sample
efficiency becomes more significant for $d > 8$.

\subsubsection{Torus}
Torus is a three-dimensional constraint introduced in Section~\ref{sec:SYMPAIS} as a
running example. We evaluate the different methods for both independent and correlated
inputs.

\begin{figure}[hbt]
    \centering
    \includegraphics[width=\linewidth]{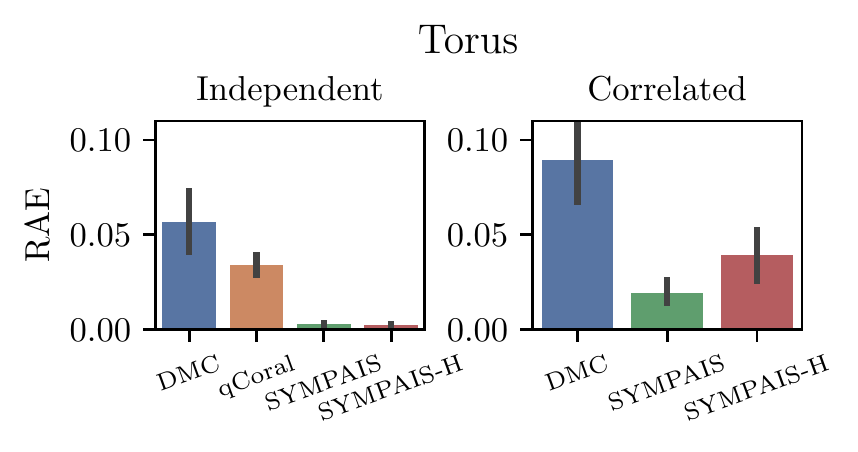}
    \caption{RAE comparison for torus.}
    \Description{RAE comparison for torus.}
    \label{fig:torus_result}
\end{figure}

\noindent\textbf{Independent inputs.} We first consider the uncorrelated input
distribution $p(\vx) = \gauss(\mathbf{0}, 0.5\mathbf{I})$ with input domain $\vx \in
[-5,5]^3 \cap \mathbb{R}^3$.
\Cref{fig:torus_result} (left) shows the RAE performance of the four methods. While
performing marginally better than the baseline DMC, \texttt{qCoral} achieve poor
performance on this non-convex subject because RealPaver fails to effectively prune out
the inner empty region of the domain within the torus, effectively reducing
\texttt{qCoral} to a DMC sampling over most of the input domain. 

RealPaver can be fine-tuned for a torus constraints by using different consistency
configurations (see~\citep{granvilliers2006RealPaver} for instructions on the matter).
However, this may require human ingenuity to select and tune the correct settings.
Finally, we observed the performance of RealPaver varies for equivalent formulations of
the constraint (e.g., $x^2$ vs $x \times x$ or reformulating the constraint without
$sqrt(\cdot)$). We conjecture that using different interval constraint propagation
algorithms or clever simplifications of the constraint may improve the performance of
\texttt{qCoral} for this problem. Both variations of SYMPAIS achieve an order of
magnitude lower RAE.

\noindent\textbf{Correlated inputs.}
Consider the correlated distribution $$p(x, y, z) = \mathcal  {T}_2(0, 0.5) \gauss(x,
0.5) \gauss(x, 0.5),$$ where ${T}_2$ denotes a Student's T distribution with 2 degrees
of freedom. Similarly to the situation illustrated
in~\cref{fig:torus-correlated-optimal-proposal}, the distribution constrained within the
solution space of torus is bi-modal. In this case, the input distribution is correlated,
with $y$ and $z$ probabilistically dependent on $x$.

Correlated and potentially multimodal input distributions are commonly used to describe
real-world inputs arising from physical phenomena.
More recently, the success of deep learning has encouraged incorporating deep neural
networks for generative modeling of high-dimensional data
distributions~\citep{rezende2015Variational,rezende2014Stochastic,kingma2014AutoEncoding},
trained from observed data. For these distributions, the PDFs are often tractable, while
CDFs often are not. This in turn makes the stratified sampling and truncations of the
input distribution intractable for \texttt{qCoral}.

On the other hand, SYMPAIS can handle these problems because it only requires evaluating
the PDF of the input distribution, not its CDF. Since \texttt{qCoral} cannot handle
correlated inputs, \cref{fig:torus_result} (right) shows how both SYMPAIS and SYMPAIS-H
outperforms the baseline DMC by nearly one order of magnitude. As discussed in
Section~\ref{sec:optimization}, seeding the MCMC processes with re-sampled diverse
solutions from the constraint solver improves SYMPAIS(-H)'s convergence for multimodal
distributions as in this experiment.

\subsection{ACAS Xu}
\begin{figure}[t]
\centering
    \includegraphics[width=\columnwidth]{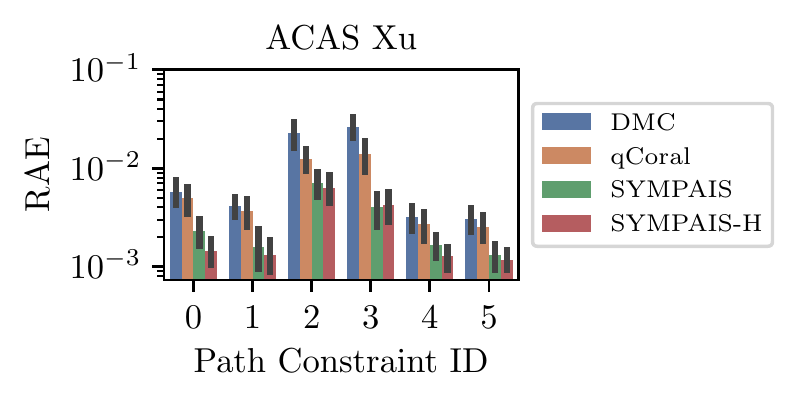}
    \caption{RAE comparison for ACAS Xu activation patterns.}
    \Description{RAE comparison for ACAS Xu activation patterns.}
    \label{fig:acasxu_result}
\end{figure}
ACAS Xu~\citep{julian2016Policy} is a benchmark neural network implementing a
safety-critical collision avoi\-dance system for unmanned aircraft control. Its inputs
are readings from a set of sensors, including: distance from the other vehicle, angle of
the other vehicles relative to ownship direction, heading angle of other vehicle, speed
of ownship, and speed of the other vehicle. The outputs of the networks are either
clear-of-conflict -- no risk of collision between ownship and the other vehicle -- or
one of four possible collision avoidance maneuvers the ownship can take to avoid a
collision. The US Federal Aviation Administration is experimenting with an
implementation of ACAS Xu to evaluate its safety for replacing the current rule-based
system~\citep{converse2020Probabilistic}.

This subject is has been used to benchmark several verification methods
(\eg~\citep{katz2017Reluplex}), including performing a probabilistic robustness analysis
that computes bounds on the probability of the network producing inconsistent decisions
for small perturbations on the inputs~\citep{converse2020Probabilistic}.
A central component of the analysis method in~\citep{converse2020Probabilistic} consists
of computing reliable bounds for the probability of satisfying a constraint that
corresponds to a unique activation pattern of the network when white noise is added to
an initial sensor reading. 

For this experiment, we extract the constraints corresponding to six random activation
patterns and estimate their satisfaction probability with different methods.

Consider a neural network with one hidden layer of $m$ neurons that receives input $\vx
\in \mathbb{R}^d$. The neural network computes the output as
\begin{align*}
    \mathbf{z} = \mathbf{W}_0 \vx + \mathbf{b}_0, \;
    \mathbf{a} = \relu(\mathbf{z}), \;
    \mathbf{y} = \mathbf{W}_1^T \mathbf{a} + \mathbf{b}_1,
\end{align*}
where $\relu$ is the Rectified Linear Unit defined as $\relu(\vx) = \max(0, \vx)$
evaluated component-wise on $\vx$. $\mathbf{W}_0, \mathbf{W}_1, \mathbf{b}_0$ and $
\mathbf{b}_1$ are the pre-trained weights of the neural network. A hidden unit
$\mathbf{a}_i$ is active if the constraint $\mathbf{z}_i \ge 0$ is satisfied and
inactive otherwise (\ie $\mathbf{z}_i < 0$). An activation pattern is defined as the
conjunction of the activation  constraints of the hidden units
$\left\{\mathbf{a}_i\right\}_{i=1}^m$. We select the network with one hidden layer of
five neurons for analysis (\url{https://bit.ly/3fjAlOW}). The selected network generates
32 possible combinations of activation patterns and we select randomly six activation
patterns for analysis. We use $\gauss(0, 1)$ to model the distribution of each input
dimension $x_i$ of the neural network and additionally impose a domain of $[-100, 100]
\cap \mathbb{R}$ for each dimension. The bounded input and independent constraints allow
the use of \texttt{qCoral} as well. However, neural networks tend to generate
high-dimensional problems because they establish control dependencies among all their
inputs, which prevents effective constraint slicing.
\begin{figure*}[bt]
    \centering
    \includegraphics{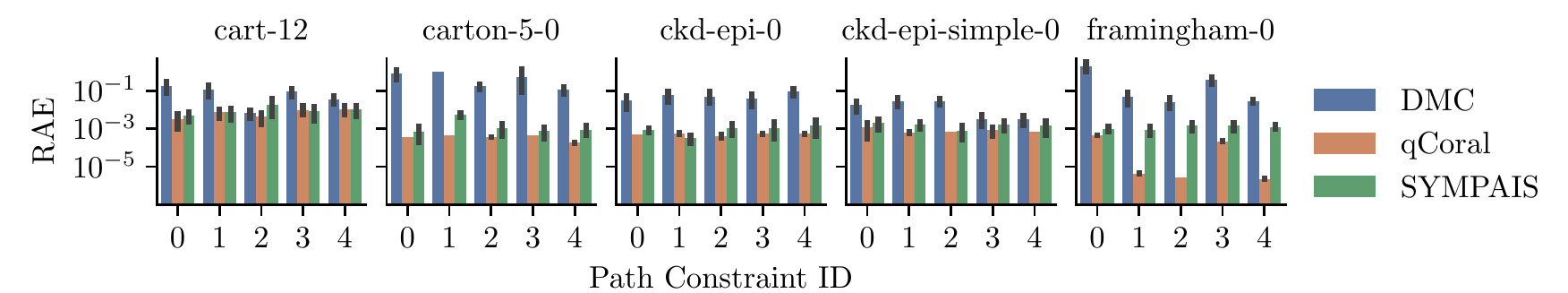}
    \caption{RAE achieved by the different methods on \texttt{volComp} subjects.}
    \Description{RAE achieved by the different methods on \texttt{volComp} subjects.}
    \label{fig:coral_result}
\end{figure*}

\Cref{fig:acasxu_result} reports the RAE achieved by the different estimation methods
for each of the six randomly sampled activation patterns. For this experiment, we used a
single initial solution computed by Z3 to initialize the MCMC chains of SYMPAIS and
SYMPAIS-H. Being the conjunction of ReLU activations, the constraints produced by ACAS
Xu are convex (intersection of half-planes) and do not induce multiple modes on the
constrained input distribution. Already using a single initial solution, SYMPAIS
converged to better estimates than DMC and \texttt{qCoral} with the same sampling
budget.

\subsection{\texttt{volComp}} Finally, in this section we experiment with a set of
constraints from the benchmark \texttt{volComp}~\citep{sankaranarayanan2013Static}, also
used to evaluate \texttt{qCoral}~\citep{borges2014Compositional,borges2015Iterative}. We
picked the first five path conditions for each of the subjects named in
\cref{fig:coral_result} from the public \texttt{qCoral} replication package. Because
most of the input variables in these subjects are computationally independent,
constraint slicing would reduce to constraints with dimensionality $<3$; we instead skip
slicing and evaluate the original constraints having between 5 and 25 variables. The
constraints are linear, with convex solution spaces. In this situation, RealPaver can
produce a tight approximation of the solution space, with significant benefits for
\texttt{qCoral}'s stratified sampling efficiency. The input CDF can be computed
analytically (independent truncated Gaussian from \texttt{qCoral}'s replication package
``normal''). Among the different experiments in~\citep{borges2014Compositional}, these
subjects represent a sweet spot for the stratified sampling method of \texttt{qCoral}
and are included here as SYMPAIS's worst-case comparison scenario.

Our current implementation of the HMC kernel proposal does not support JIT-compilable
truncated distributions. Thus we run only SYMPAIS with RWMH for this set of subjects.

\cref{fig:coral_result} shows the RAE of the different methods. The ground truth is
computed with Mathematica. For all the subjects in this experiment, except
\texttt{cart-12}, $99\%$ of the input domain enclosed within RealPaver's boxes contains
only solutions of the constraint. For all the subjects, except \texttt{framingham-0},
\texttt{qCoral} and SYMPAIS produce comparable RAE. A deeper inspection of
\texttt{framingham-0} showed that most of the constraint is effectively the intersection
of boxes, which are identified as inner boxes by RealPaver and require no further
sampling for probability estimation (Equation~\eqref{eqn:stratified}). The experiment
demonstrates that while the adaptive importance sampling strategy of SYMPAIS is designed
to estimate the satisfaction probability of high-dimensional constraints with
multimodal, correlated input distributions, it can match the performance of stratified
sampling for most simpler problems where stratified sampling can be applied. The
analysis of the same constraints with correlated inputs would instead not be possible
with stratified sampling.

%% file: sections/related_work.tex
\section{Related Work}

Probabilistic symbolic execution~\citep{geldenhuys2012Probabilistic} relies on symbolic
execution to extract the path conditions characterizing the inputs that lead to the
occurrence of a target event; the probability of satisfying the path condition
constraints given an input distribution is then quantified using model counting or
solution space quantification methods. PSE has been applied in several domains,
including reliability~\citep{filieri2013Reliability}
security~\citep{malacaria2018Symbolic,brennan2018Symbolic} and performance
analyses~\citep{chen2016Generating}, with variations implemented for
nondeterministic~\citep{luckow2014Exact} and probabilistic~\citep{malacaria2018Symbolic}
programs. Quantification methods have been proposed for uniform or discretized
distributions over linear integer constraints~\citep{filieri2013Reliability}, string
constraints~\citep{aydin2018Parameterized}, bounded data
structured~\citep{filieri2015Model}, and numerical constraints over continuous input
distributions~\cite{borges2014Compositional,borges2015Iterative}. 

In the probabilistic programming literature, \citet{chaganty2013Efficiently} proposed
breaking a probabilistic program with branches and loops into small programs focused on
only some execution paths and use pre-image analysis to perform efficient importance
sampling. Differently from~\citet{chaganty2013Efficiently}, we use PI-MAIS and MCMC
processes to further adapt the proposal distributions for the analysis of individual
path conditions. \citet{nori2014R2} similarly uses the idea of pre-image analysis to
design a proposal distribution that generates samples that are less likely to be
rejected in MCMC. These analyses complement our approach and can potentially be
incorporated to improve our MCMC scheme. Recent work by~\citet{zhou2020Divide} motivates
the decomposition into subproblems by considering universal probabilistic programs with
stochastic support, \ie depending on the values of the samples, the program may take on
different control-flow paths, and the number of random variables evaluated along each
path varies as a result. This makes designing a proposal distribution for efficient MCMC
difficult. \citet{zhou2020Divide} approaches this issue by decomposing the problem
into small straight-line programs (SLPs) for which the support is fixed and posterior
inference is tractable. However, differently from PSE approaches, SLPs are execution paths
sampled via a specialized MCMC algorithm, which adds an additional degree of uncertainty
to the results of probabilistic analysis and is not suitable for the analysis of rare events.

%% file: sections/conclusions.tex
\section{Conclusions and Future Work}\label{secConclusions} We introduced SYMPAIS, a new
inference method for estimating the satisfaction probability of numerical constraints on
high-dimensional, correlated input distributions. SYMPAIS combines a sample-efficient
importance sampling scheme with constraint solvers to extend the applicability of
probabilistic symbolic execution to a broader class of programs processing correlated
inputs that cannot be analyzed with existing methods. 
While we currently implemented only RWMH and HMC kernel as adaptive proposals, SYMPAIS
can be extended with additional kernels to improve its performance on different classes
of constraints.
Finally, it is also worth investigating the integration of kernels and
parametric importance sampling proposals for discrete distributions, aiming at
supporting integer input variables that cannot be analyzed with our current algorithms.

%% file: sections/appendix.tex
\section*{Appendix}
\section{Experimental settings details}
\noindent\textbf{Implementation.} We implemented SYMPAIS in Python 3.8 using Google's
JAX framework~\citep{bradbury2018JAX} to implement the mathematical procedures and
generate random samples. JAX provides automated differentiation, which we use for HMC,
and allows for just-in-time compile the numerical routines of SYMPAIS for faster
execution, mainly due to the use of its automatic vectorization features.

\noindent\textbf{Execution environment.} All the experiments have been run on an AMD
EPYC 7401P CPU with 448Gb of memory. However, SYMPAIS was limited to using only two
cores and never exceeded 4Gb of memory consumption in our experiments.

\noindent\textbf{Configuration.} The RWMH kernel is configured with scale $\sigma =
1.0$. The HMC trajectory is simulated for $T=20$ steps, with each time step duration
$\epsilon=0.1$; we use standard leapfrog steps to generate the Hamiltonian
trajectory~\citep{neal2011MCMC} (mathematical formulation in the Appendix). We use
$N=100$ parallel MCMC chains. Each chain is warmed up with $500$ samples -- \ie $500$
samples are used to bring the chain near convergence and discarded; only the samples
from the $501-th$ on are actually used. Overall, the $100$ chains take $50,000$ samples
to warm up. For the RWMH kernel, we also implement adaptation of the scale parameter
following in scheme used by PyMC3 (\url{https://bit.ly/2HrpHJK}).

For the adaptive importance sampling distribution $q_{n, t}(\cdot)$, we use
$\gauss(\cdot | \vx{n,t}, 0.5 \Lambda)$ where $\Lambda$ is the covariance of the input
distribution $p(\vx)$. For the input distributions where the covariance of $p(\vx)$ is
analytically intractable, SYMPAIS uses the sample variance (estimated from $100$
samples). We draw $M=5$ samples from each sub-proposal $q_{n,t}$ during each iteration
of the adaptation loop. RealPaver is configured with accuracy $0.1$ to generate the
initial inputs of the MCMC chains and the truncated kernel proposal bounds (min/max over
RealPaver's boxes vertices). RealPaver for paving in \texttt{qCoral} computes up to 1024
boxes.

\noindent\textbf{Experimental settings.} We allow each of the four methods the same
budget of $10^6$ samples. For SYMPAIS, this includes the samples used to warm up the
MCMC chains and to estimate the covariance of the input distribution when not given.
RealPaver executions for qCoral and SYMPAIS are not time-limited. However, all the
experiments completed their execution within 2 minutes for the geometrical
microbenchmarks and 5 minutes for the other subjects. The compilation time of JAX varies
between $\tilde 20$ seconds and $\tilde 25$ minutes; reimplementing SYMPAIS with a
compiled language might make the compilation time more predictable. 

Each experiment is repeated $20$ times unless otherwise specified. To compare the
estimates of the different methods, we use the relative absolute error (RAE) against a
ground truth value computed with Mathematica (accuracy and precision goals = 15) or via
a large direct Monte Carlo samples ($20\cdot 10^{8}$) for the geometrical
microbenchmarks. The RAE is preferred to the absolute error because normalizing by the
ground truth probability allows a homogenous comparison over constraints with different
satisfaction probabilities.

\section{Hamiltonian Monte Carlo estimators}\label{appendixHMC}
The Hamiltonian Markov Chain Monte Carlo (HMC)
algorithm~\citep{neal2011MCMC,betancourt2018Conceptual} uses derivatives of the target
distribution and auxiliary variables to generate candidate samples more efficiently than
RWMH when $p(\vx)$ is differentiable.

HMC generates the next sample $\vx'$ by simulating the movement of a particle pushed in
a random direction starting from the location $\vx$ across the surface defined by the
probability density function of $p(\vx)$. Intuitively, regions where $p(\vx)$ is larger
are at a ``lower altitude'' on the surface and are thus more likely to attract the
particle. The random direction and energy of the push can allow the particle to reach
any point of the surface. The position of the particle after a predefined fixed time $T$
will be the new candidate sample $\vx'$. A mathematical formulation of the HMC process
implemented in SYMPAIS is reported in the Appendix, while for a broader presentation of
the method, the reader is referred to~\citep{betancourt2018Conceptual}.

To sample from the target distrbution of $\vx$, HMC augments the original probability
distribution \(\overline{p}(\vx)\) with auxiliary momentum variables
\(\boldsymbol{\rho}\) and considers the joint density \(\overline{p}(\vx,
\boldsymbol{\rho}) = p(\boldsymbol{\rho})\overline{p}(\vx)\) where
\(p(\boldsymbol{\rho}\nolinebreak\mid\vx)\) can be chosen as \(\gauss(\mathbf{0},
\mathbf{I})\), $\mathbf{I}$ being the identity matrix with size $|\vx|$. Define the
Hamiltonian \(H\) as
\begin{align*}
    H(\vx, \boldsymbol{\rho}) &= -\log \overline{p} (\vx, \boldsymbol{\rho}) \\
    &= -\log p(\boldsymbol{\rho}) - \log \overline{p}(\vx) \\
    &= K(\boldsymbol{\rho}) + U(\vx)
\end{align*}
where \(K(\boldsymbol{\rho}) = -\log p(\boldsymbol{\rho})\) is called the \emph{kinetic
energy} and \(U(\vx) = -\log\overline{p}(\vx)\) is called the \emph{potential energy}.

A proposal $\vx'$ is made by simulating the Hamiltonian dynamics
\[
    \frac{d \vx}{d t} = \nabla_{\boldsymbol{\rho}}K(\boldsymbol{\rho}), \;
    \frac{d \boldsymbol{\rho}}{d t} = -\nabla_{\vx}U(\vx),
\]
which can be discretized and solved numerically by a symplectic integrator such as the
Leapfrog method for \(T_{\text{leapfrog}}\) steps with a step size of \(\epsilon\)
\begin{align*}
    \boldsymbol{\rho}_{t + \epsilon/2} &= \boldsymbol{\rho}_{t} - (\epsilon/2)\nabla_{\vx}U(\vx_t), \\
    \vx_{t + \epsilon} &= \vx_{t} + \epsilon \nabla_{\boldsymbol{\rho}}K(\boldsymbol{\rho}_{t+\epsilon/2}), \\
    \boldsymbol{\rho}_{t + \epsilon} &= \boldsymbol{\rho}_{t + \epsilon/2} - (\epsilon/2)\nabla_{\vx}U(\vx_{t + \epsilon})
\end{align*}
for \(t \in \{1,\dots,T_{\text{leapfrog}}\}\).

Choice of hyperparameters parameters such as the step size \(\epsilon\) and the number
of Leapfrog steps \(T_{\text{leapfrog}}\) affects the efficiency of HMC methods. The
No-U-Turn Sampler ~\citep{hoffman2014NUTs} is an extension to the HMC method which
mitigates the effect of hyperparameter settings on HMC methods.

While potentially more efficient than RWMH, especially for higher-dimensional problems,
-- \ie fewer candidate samples are rejected -- HMC requires the target density function
to be differentiable, which makes it less general than RWMH. In particular, it cannot be
applied for input distribution with discrete parameters. Improvements to the original
HMC algorithm and extensions exist to handle this
limitation~\citep{nishimura2020Discontinuous}. Because SYMPAIS's target distribution is
$\bar{p}(\vx)=p(\vx)\ind_{C}$, instead of the original input distribution $p(\vx)$, the
probability density is not differentiable at the boundary of the solution space of the
constraint $C$, and simulations crossing that boundary have to be rejected. Extensions
to HMC specialized for sampling from constrained spaces have been proposed for several
classes of constraints and will be investigated for future extensions of
SYMPAIS~\citep{afshar2015Reflection,betancourt2018Conceptual,nishimura2020Discontinuous,neal2011MCMC}.

\section{Additional discussion}

\noindent\textbf{Handling broader families of $p(\vx)$.} Since SYMPAIS does not require
CDF of $p(\vx)$ to be tractable, we only require that the PDF can be evaluated. This
allows us to handle a broader family of distributions compared to qCoral. This is a
trade-off between efficiency and generality. By assuming that the truncated CDF can be
evaluated. qCoral can perform additional pruning that further eliminates probability
masses where $\ind_{PC}(\vx) = 0$. While stratified sampling can be used to reduce
variance~\citep{keramat1998study}, most of the variance reduction in qCoral comes from
pruning out irrelevant solution space instead of stratification. SYMPAIS, on the other
hand, relaxes the assumption of having tractable truncated CDFs but instead reduces
variance with IS. The interval contraction optimization allows us to reduce variance
compared to AIS in the original space, but this can still produce a larger truncated
proposal compared to qCoral. However, IS may provide additional variance reduction to
compensate for the inefficiency in the stratified sampler in qCoral. In this case,
SYMPAIS complements qCoral in situations where the stratification strategy is
inefficient. When the truncated CDF is available, which method works better is likely
problem-specific. In this case, the most practical solution is probably a combination of
both. One can first do a preliminary run with a smaller number of samples to identify
which method would work better and exploit the more efficient method.

\noindent\textbf{Future improvements} Improvements to the MCMC algorithm can improve
SYMPAIS's efficiency. In particular, it would benefit from improvements in MCMC that can
explore multimodal posterior efficiently. Note that, however, the MCMC does not have to
be perfect. For the PI-MAIS adaptation scheme, the crucial thing is that the parallel
chains should capture the modes in $q^*(\vx)$ in a way the modes are relatively weighted
in $q^*(\vx)$. This may explain why we do not observe improvements in performance when
using the HMC kernel. Importance sampling can be dangerous if the adapted $q$ misses the
modes of $q^*$; this is partly mitigated in SYMPAIS, which uses an adaptive mixture
proposal that can approximate multimodal posteriors. Additionally, we plan to further
improve the robustness of SYMPAIS against multimodal distributions adding supplementary
diagnostics (\eg Effective sample sizes (ESS)) and defensive importance
sampling~\citep{owen2013Monte,hesterberg1995Weighted}.

%% file: main.bbl

\begin{thebibliography}{48}


\ifx \showCODEN    \undefined \def \showCODEN     #1{\unskip}     \fi
\ifx \showDOI      \undefined \def \showDOI       #1{#1}\fi
\ifx \showISBNx    \undefined \def \showISBNx     #1{\unskip}     \fi
\ifx \showISBNxiii \undefined \def \showISBNxiii  #1{\unskip}     \fi
\ifx \showISSN     \undefined \def \showISSN      #1{\unskip}     \fi
\ifx \showLCCN     \undefined \def \showLCCN      #1{\unskip}     \fi
\ifx \shownote     \undefined \def \shownote      #1{#1}          \fi
\ifx \showarticletitle \undefined \def \showarticletitle #1{#1}   \fi
\ifx \showURL      \undefined \def \showURL       {\relax}        \fi
\providecommand\bibfield[2]{#2}
\providecommand\bibinfo[2]{#2}
\providecommand\natexlab[1]{#1}
\providecommand\showeprint[2][]{arXiv:#2}

\bibitem[\protect\citeauthoryear{Afshar and Domke}{Afshar and Domke}{2015}]%
        {afshar2015Reflection}
\bibfield{author}{\bibinfo{person}{Hadi~Mohasel Afshar} {and}
  \bibinfo{person}{Justin Domke}.} \bibinfo{year}{2015}\natexlab{}.
\newblock \showarticletitle{Reflection, Refraction, and {{Hamiltonian Monte
  Carlo}}}. In \bibinfo{booktitle}{\emph{Proceedings of the 28th
  {{International Conference}} on {{Neural Information Processing Systems}} -
  {{Volume}} 2}} \emph{(\bibinfo{series}{{{NIPS}}'15})}.
  \bibinfo{publisher}{{MIT Press}}, \bibinfo{pages}{3007--3015}.
\newblock


\bibitem[\protect\citeauthoryear{Araya, Trombettoni, and Neveu}{Araya
  et~al\mbox{.}}{2012}]%
        {araya2012Contractor}
\bibfield{author}{\bibinfo{person}{Ignacio Araya}, \bibinfo{person}{Gilles
  Trombettoni}, {and} \bibinfo{person}{Bertrand Neveu}.}
  \bibinfo{year}{2012}\natexlab{}.
\newblock \showarticletitle{A {{Contractor Based}} on {{Convex Interval
  Taylor}}}. In \bibinfo{booktitle}{\emph{Integration of {{AI}} and {{OR
  Techniques}} in {{Contraint Programming}} for {{Combinatorial Optimzation
  Problems}}}} \emph{(\bibinfo{series}{Lecture {{Notes}} in {{Computer
  Science}}})}, \bibfield{editor}{\bibinfo{person}{Nicolas Beldiceanu},
  \bibinfo{person}{Narendra Jussien}, {and} \bibinfo{person}{{\'E}ric Pinson}}
  (Eds.). \bibinfo{publisher}{{Springer}}, \bibinfo{pages}{1--16}.
\newblock
\showISBNx{978-3-642-29828-8}
\urldef\tempurl%
\url{https://doi.org/10.1007/978-3-642-29828-8_1}
\showDOI{\tempurl}


\bibitem[\protect\citeauthoryear{Aydin, Eiers, Bang, Brennan, Gavrilov, Bultan,
  and Yu}{Aydin et~al\mbox{.}}{2018}]%
        {aydin2018Parameterized}
\bibfield{author}{\bibinfo{person}{Abdulbaki Aydin}, \bibinfo{person}{William
  Eiers}, \bibinfo{person}{Lucas Bang}, \bibinfo{person}{Tegan Brennan},
  \bibinfo{person}{Miroslav Gavrilov}, \bibinfo{person}{Tevfik Bultan}, {and}
  \bibinfo{person}{Fang Yu}.} \bibinfo{year}{2018}\natexlab{}.
\newblock \showarticletitle{Parameterized Model Counting for String and Numeric
  Constraints}. In \bibinfo{booktitle}{\emph{Proceedings of the 2018 26th {{ACM
  Joint Meeting}} on {{European Software Engineering Conference}} and
  {{Symposium}} on the {{Foundations}} of {{Software Engineering}}}}.
  \bibinfo{publisher}{{ACM}}, \bibinfo{pages}{400--410}.
\newblock
\showISBNx{978-1-4503-5573-5}
\urldef\tempurl%
\url{https://doi.org/10.1145/3236024.3236064}
\showDOI{\tempurl}


\bibitem[\protect\citeauthoryear{Betancourt}{Betancourt}{2018}]%
        {betancourt2018Conceptual}
\bibfield{author}{\bibinfo{person}{Michael Betancourt}.}
  \bibinfo{year}{2018}\natexlab{}.
\newblock \showarticletitle{A {{Conceptual Introduction}} to {{Hamiltonian
  Monte Carlo}}}.
\newblock \bibinfo{journal}{\emph{arXiv:1701.02434 [stat]}}
  (\bibinfo{date}{July} \bibinfo{year}{2018}).
\newblock
\showeprint[arxiv]{1701.02434}~[stat]


\bibitem[\protect\citeauthoryear{Borges, Filieri, D'Amorim, and P{\u a}s{\u
  a}reanu}{Borges et~al\mbox{.}}{2015}]%
        {borges2015Iterative}
\bibfield{author}{\bibinfo{person}{Mateus Borges}, \bibinfo{person}{Antonio
  Filieri}, \bibinfo{person}{Marcelo D'Amorim}, {and}
  \bibinfo{person}{Corina~S. P{\u a}s{\u a}reanu}.}
  \bibinfo{year}{2015}\natexlab{}.
\newblock \showarticletitle{Iterative Distribution-Aware Sampling for
  Probabilistic Symbolic Execution}. In \bibinfo{booktitle}{\emph{Proceedings
  of the 2015 10th {{Joint Meeting}} on {{Foundations}} of {{Software
  Engineering}}}}. \bibinfo{publisher}{{ACM}}, \bibinfo{pages}{866--877}.
\newblock
\showISBNx{978-1-4503-3675-8}
\urldef\tempurl%
\url{https://doi.org/10.1145/2786805.2786832}
\showDOI{\tempurl}


\bibitem[\protect\citeauthoryear{Borges, Filieri, {d'Amorim}, P{\u a}s{\u
  a}reanu, and Visser}{Borges et~al\mbox{.}}{2014}]%
        {borges2014Compositional}
\bibfield{author}{\bibinfo{person}{Mateus Borges}, \bibinfo{person}{Antonio
  Filieri}, \bibinfo{person}{Marcelo {d'Amorim}}, \bibinfo{person}{Corina~S.
  P{\u a}s{\u a}reanu}, {and} \bibinfo{person}{Willem Visser}.}
  \bibinfo{year}{2014}\natexlab{}.
\newblock \showarticletitle{Compositional Solution Space Quantification for
  Probabilistic Software Analysis}. In \bibinfo{booktitle}{\emph{Proceedings of
  the 35th {{ACM SIGPLAN Conference}} on {{Programming Language Design}} and
  {{Implementation}}}}. \bibinfo{publisher}{{ACM}}, \bibinfo{pages}{123--132}.
\newblock
\showISBNx{978-1-4503-2784-8}
\urldef\tempurl%
\url{https://doi.org/10.1145/2594291.2594329}
\showDOI{\tempurl}


\bibitem[\protect\citeauthoryear{Bradbury, Frostig, Hawkins, Johnson, Leary,
  Maclaurin, Necula, Paszke, VanderPlas, {Wanderman-Milne}, and Zhang}{Bradbury
  et~al\mbox{.}}{2018}]%
        {bradbury2018JAX}
\bibfield{author}{\bibinfo{person}{James Bradbury}, \bibinfo{person}{Roy
  Frostig}, \bibinfo{person}{Peter Hawkins}, \bibinfo{person}{Matthew~James
  Johnson}, \bibinfo{person}{Chris Leary}, \bibinfo{person}{Dougal Maclaurin},
  \bibinfo{person}{George Necula}, \bibinfo{person}{Adam Paszke},
  \bibinfo{person}{Jake VanderPlas}, \bibinfo{person}{Skye {Wanderman-Milne}},
  {and} \bibinfo{person}{Qiao Zhang}.} \bibinfo{year}{2018}\natexlab{}.
\newblock \bibinfo{title}{{{JAX}}: Composable Transformations of
  {{Python}}+{{NumPy}} Programs}.
\newblock
\newblock
\urldef\tempurl%
\url{https://github.com/google/jax}
\showURL{%
\tempurl}


\bibitem[\protect\citeauthoryear{Brennan, Saha, Bultan, and P{\u a}s{\u
  a}reanu}{Brennan et~al\mbox{.}}{2018}]%
        {brennan2018Symbolic}
\bibfield{author}{\bibinfo{person}{Tegan Brennan}, \bibinfo{person}{Seemanta
  Saha}, \bibinfo{person}{Tevfik Bultan}, {and} \bibinfo{person}{Corina~S. P{\u
  a}s{\u a}reanu}.} \bibinfo{year}{2018}\natexlab{}.
\newblock \showarticletitle{Symbolic Path Cost Analysis for Side-Channel
  Detection}. In \bibinfo{booktitle}{\emph{Proceedings of the 27th {{ACM
  SIGSOFT International Symposium}} on {{Software Testing}} and {{Analysis}}}}.
  \bibinfo{publisher}{{ACM}}, \bibinfo{pages}{27--37}.
\newblock
\showISBNx{978-1-4503-5699-2}
\urldef\tempurl%
\url{https://doi.org/10.1145/3213846.3213867}
\showDOI{\tempurl}


\bibitem[\protect\citeauthoryear{B{\"u}eler, Enge, and Fukuda}{B{\"u}eler
  et~al\mbox{.}}{2000}]%
        {bueler2000Exact}
\bibfield{author}{\bibinfo{person}{Benno B{\"u}eler}, \bibinfo{person}{Andreas
  Enge}, {and} \bibinfo{person}{Komei Fukuda}.}
  \bibinfo{year}{2000}\natexlab{}.
\newblock \showarticletitle{Exact {{Volume Computation}} for {{Polytopes}}: {{A
  Practical Study}}}.
\newblock In \bibinfo{booktitle}{\emph{Polytopes \textemdash{}
  {{Combinatorics}} and {{Computation}}}},
  \bibfield{editor}{\bibinfo{person}{Gil Kalai} {and}
  \bibinfo{person}{G{\"u}nter~M. Ziegler}} (Eds.).
  \bibinfo{publisher}{{Birkh\"auser Basel}}, \bibinfo{pages}{131--154}.
\newblock
\showISBNx{978-3-7643-6351-2 978-3-0348-8438-9}
\urldef\tempurl%
\url{https://doi.org/10.1007/978-3-0348-8438-9_6}
\showDOI{\tempurl}


\bibitem[\protect\citeauthoryear{Cadar, Dunbar, and Engler}{Cadar
  et~al\mbox{.}}{2008}]%
        {cadar2008KLEE}
\bibfield{author}{\bibinfo{person}{Cristian Cadar}, \bibinfo{person}{Daniel
  Dunbar}, {and} \bibinfo{person}{Dawson Engler}.}
  \bibinfo{year}{2008}\natexlab{}.
\newblock \showarticletitle{{{KLEE}}: Unassisted and Automatic Generation of
  High-Coverage Tests for Complex Systems Programs}. In
  \bibinfo{booktitle}{\emph{Proceedings of the 8th {{USENIX}} Conference on
  {{Operating}} Systems Design and Implementation}}
  \emph{(\bibinfo{series}{{{OSDI}}'08})}. \bibinfo{publisher}{{USENIX
  Association}}, \bibinfo{pages}{209--224}.
\newblock


\bibitem[\protect\citeauthoryear{Chaganty, Nori, and Rajamani}{Chaganty
  et~al\mbox{.}}{2013}]%
        {chaganty2013Efficiently}
\bibfield{author}{\bibinfo{person}{Arun Chaganty}, \bibinfo{person}{Aditya
  Nori}, {and} \bibinfo{person}{Sriram Rajamani}.}
  \bibinfo{year}{2013}\natexlab{}.
\newblock \showarticletitle{Efficiently Sampling Probabilistic Programs via
  Program Analysis}. In \bibinfo{booktitle}{\emph{Proceedings of the Sixteenth
  International Conference on Artificial Intelligence and Statistics}}
  \emph{(\bibinfo{series}{Proceedings of Machine Learning Research},
  Vol.~\bibinfo{volume}{31})}, \bibfield{editor}{\bibinfo{person}{Carlos~M.
  Carvalho} {and} \bibinfo{person}{Pradeep Ravikumar}} (Eds.).
  \bibinfo{publisher}{{PMLR}}, \bibinfo{pages}{153--160}.
\newblock


\bibitem[\protect\citeauthoryear{Chen, Liu, and Le}{Chen et~al\mbox{.}}{2016}]%
        {chen2016Generating}
\bibfield{author}{\bibinfo{person}{Bihuan Chen}, \bibinfo{person}{Yang Liu},
  {and} \bibinfo{person}{Wei Le}.} \bibinfo{year}{2016}\natexlab{}.
\newblock \showarticletitle{Generating Performance Distributions via
  Probabilistic Symbolic Execution}. In \bibinfo{booktitle}{\emph{Proceedings
  of the 38th {{International Conference}} on {{Software Engineering}}}}.
  \bibinfo{publisher}{{ACM}}, \bibinfo{pages}{49--60}.
\newblock
\showISBNx{978-1-4503-3900-1}
\urldef\tempurl%
\url{https://doi.org/10.1145/2884781.2884794}
\showDOI{\tempurl}


\bibitem[\protect\citeauthoryear{Converse, Filieri, Gopinath, and
  Pasareanu}{Converse et~al\mbox{.}}{2020}]%
        {converse2020Probabilistic}
\bibfield{author}{\bibinfo{person}{Hayes Converse}, \bibinfo{person}{Antonio
  Filieri}, \bibinfo{person}{Divya Gopinath}, {and} \bibinfo{person}{Corina~S.
  Pasareanu}.} \bibinfo{year}{2020}\natexlab{}.
\newblock \showarticletitle{Probabilistic {{Symbolic Analysis}} of {{Neural
  Networks}}}. In \bibinfo{booktitle}{\emph{2020 {{IEEE}} 31st {{International
  Symposium}} on {{Software Reliability Engineering}} ({{ISSRE}})}}.
  \bibinfo{publisher}{{IEEE}}, \bibinfo{pages}{148--159}.
\newblock
\showISBNx{978-1-72819-870-5}
\urldef\tempurl%
\url{https://doi.org/10.1109/ISSRE5003.2020.00023}
\showDOI{\tempurl}


\bibitem[\protect\citeauthoryear{{de Moura} and Bj{\o}rner}{{de Moura} and
  Bj{\o}rner}{2008}]%
        {demoura2008Z3}
\bibfield{author}{\bibinfo{person}{Leonardo {de Moura}} {and}
  \bibinfo{person}{Nikolaj Bj{\o}rner}.} \bibinfo{year}{2008}\natexlab{}.
\newblock \showarticletitle{Z3: {{An Efficient SMT Solver}}}. In
  \bibinfo{booktitle}{\emph{Tools and {{Algorithms}} for the {{Construction}}
  and {{Analysis}} of {{Systems}}}} \emph{(\bibinfo{series}{Lecture {{Notes}}
  in {{Computer Science}}})}, \bibfield{editor}{\bibinfo{person}{C.~R.
  Ramakrishnan} {and} \bibinfo{person}{Jakob Rehof}} (Eds.).
  \bibinfo{publisher}{{Springer}}, \bibinfo{pages}{337--340}.
\newblock
\showISBNx{978-3-540-78800-3}
\urldef\tempurl%
\url{https://doi.org/10.1007/978-3-540-78800-3_24}
\showDOI{\tempurl}


\bibitem[\protect\citeauthoryear{Dwyer, Filieri, Geldenhuys, Gerrard, P{\u
  a}s{\u a}reanu, and Visser}{Dwyer et~al\mbox{.}}{2017}]%
        {dwyer2017Probabilistic}
\bibfield{author}{\bibinfo{person}{Matthew~B. Dwyer}, \bibinfo{person}{Antonio
  Filieri}, \bibinfo{person}{Jaco Geldenhuys}, \bibinfo{person}{Mitchell
  Gerrard}, \bibinfo{person}{Corina~S. P{\u a}s{\u a}reanu}, {and}
  \bibinfo{person}{Willem Visser}.} \bibinfo{year}{2017}\natexlab{}.
\newblock \showarticletitle{Probabilistic {{Program Analysis}}}.
\newblock In \bibinfo{booktitle}{\emph{Grand {{Timely Topics}} in {{Software
  Engineering}}}}, \bibfield{editor}{\bibinfo{person}{J{\'a}come Cunha},
  \bibinfo{person}{Jo{\~a}o~P. Fernandes}, \bibinfo{person}{Ralf L{\"a}mmel},
  \bibinfo{person}{Jo{\~a}o Saraiva}, {and} \bibinfo{person}{Vadim Zaytsev}}
  (Eds.). Vol.~\bibinfo{volume}{10223}. \bibinfo{publisher}{{Springer
  International Publishing}}, \bibinfo{pages}{1--25}.
\newblock
\showISBNx{978-3-319-60073-4 978-3-319-60074-1}
\urldef\tempurl%
\url{https://doi.org/10.1007/978-3-319-60074-1_1}
\showDOI{\tempurl}


\bibitem[\protect\citeauthoryear{Filieri, Frias, P{\u a}s{\u a}reanu, and
  Visser}{Filieri et~al\mbox{.}}{2015}]%
        {filieri2015Model}
\bibfield{author}{\bibinfo{person}{Antonio Filieri},
  \bibinfo{person}{Marcelo~F. Frias}, \bibinfo{person}{Corina~S. P{\u a}s{\u
  a}reanu}, {and} \bibinfo{person}{Willem Visser}.}
  \bibinfo{year}{2015}\natexlab{}.
\newblock \showarticletitle{Model {{Counting}} for {{Complex Data
  Structures}}}. In \bibinfo{booktitle}{\emph{Model {{Checking Software}}}}
  \emph{(\bibinfo{series}{Lecture {{Notes}} in {{Computer Science}}})},
  \bibfield{editor}{\bibinfo{person}{Bernd Fischer} {and} \bibinfo{person}{Jaco
  Geldenhuys}} (Eds.). \bibinfo{publisher}{{Springer International
  Publishing}}, \bibinfo{pages}{222--241}.
\newblock
\showISBNx{978-3-319-23404-5}
\urldef\tempurl%
\url{https://doi.org/10.1007/978-3-319-23404-5_15}
\showDOI{\tempurl}


\bibitem[\protect\citeauthoryear{Filieri, Pasareanu, and Visser}{Filieri
  et~al\mbox{.}}{2013}]%
        {filieri2013Reliability}
\bibfield{author}{\bibinfo{person}{Antonio Filieri}, \bibinfo{person}{Corina~S.
  Pasareanu}, {and} \bibinfo{person}{Willem Visser}.}
  \bibinfo{year}{2013}\natexlab{}.
\newblock \showarticletitle{Reliability Analysis in {{Symbolic PathFinder}}}.
  In \bibinfo{booktitle}{\emph{2013 35th {{International Conference}} on
  {{Software Engineering}} ({{ICSE}})}}. \bibinfo{publisher}{{IEEE}},
  \bibinfo{pages}{622--631}.
\newblock
\showISBNx{978-1-4673-3076-3 978-1-4673-3073-2}
\urldef\tempurl%
\url{https://doi.org/10.1109/ICSE.2013.6606608}
\showDOI{\tempurl}


\bibitem[\protect\citeauthoryear{Filieri, P{\u a}s{\u a}reanu, Visser, and
  Geldenhuys}{Filieri et~al\mbox{.}}{2014}]%
        {filieri2014Statistical}
\bibfield{author}{\bibinfo{person}{Antonio Filieri}, \bibinfo{person}{Corina~S.
  P{\u a}s{\u a}reanu}, \bibinfo{person}{Willem Visser}, {and}
  \bibinfo{person}{Jaco Geldenhuys}.} \bibinfo{year}{2014}\natexlab{}.
\newblock \showarticletitle{Statistical Symbolic Execution with Informed
  Sampling}. In \bibinfo{booktitle}{\emph{Proceedings of the 22nd {{ACM SIGSOFT
  International Symposium}} on {{Foundations}} of {{Software Engineering}}}}.
  \bibinfo{publisher}{{ACM}}, \bibinfo{pages}{437--448}.
\newblock
\showISBNx{978-1-4503-3056-5}
\urldef\tempurl%
\url{https://doi.org/10.1145/2635868.2635899}
\showDOI{\tempurl}


\bibitem[\protect\citeauthoryear{Gehr, Misailovic, and Vechev}{Gehr
  et~al\mbox{.}}{2016}]%
        {gehr2016PSI}
\bibfield{author}{\bibinfo{person}{Timon Gehr}, \bibinfo{person}{Sasa
  Misailovic}, {and} \bibinfo{person}{Martin Vechev}.}
  \bibinfo{year}{2016}\natexlab{}.
\newblock \showarticletitle{{{PSI}}: {{Exact Symbolic Inference}} for
  {{Probabilistic Programs}}}.
\newblock In \bibinfo{booktitle}{\emph{Computer {{Aided Verification}}}},
  \bibfield{editor}{\bibinfo{person}{Swarat Chaudhuri} {and}
  \bibinfo{person}{Azadeh Farzan}} (Eds.). Vol.~\bibinfo{volume}{9779}.
  \bibinfo{publisher}{{Springer International Publishing}},
  \bibinfo{pages}{62--83}.
\newblock
\showISBNx{978-3-319-41527-7 978-3-319-41528-4}
\urldef\tempurl%
\url{https://doi.org/10.1007/978-3-319-41528-4_4}
\showDOI{\tempurl}


\bibitem[\protect\citeauthoryear{Geldenhuys, Dwyer, and Visser}{Geldenhuys
  et~al\mbox{.}}{2012}]%
        {geldenhuys2012Probabilistic}
\bibfield{author}{\bibinfo{person}{Jaco Geldenhuys},
  \bibinfo{person}{Matthew~B. Dwyer}, {and} \bibinfo{person}{Willem Visser}.}
  \bibinfo{year}{2012}\natexlab{}.
\newblock \showarticletitle{Probabilistic Symbolic Execution}. In
  \bibinfo{booktitle}{\emph{Proceedings of the 2012 {{International Symposium}}
  on {{Software Testing}} and {{Analysis}} - {{ISSTA}} 2012}}.
  \bibinfo{publisher}{{ACM Press}}, \bibinfo{pages}{166}.
\newblock
\showISBNx{978-1-4503-1454-1}
\urldef\tempurl%
\url{https://doi.org/10.1145/2338965.2336773}
\showDOI{\tempurl}


\bibitem[\protect\citeauthoryear{Goldsztejn and Chabert}{Goldsztejn and
  Chabert}{2021}]%
        {goldsztejnibexlib}
\bibfield{author}{\bibinfo{person}{Alexandre Goldsztejn} {and}
  \bibinfo{person}{Gilles Chabert}.} \bibinfo{year}{2021}\natexlab{}.
\newblock \bibinfo{title}{{ibex-lib}}.
\newblock
\newblock
\urldef\tempurl%
\url{http://www.ibex-lib.org}
\showURL{%
Retrieved Jun 4, 2021 from \tempurl}


\bibitem[\protect\citeauthoryear{Granvilliers and Benhamou}{Granvilliers and
  Benhamou}{2006}]%
        {granvilliers2006RealPaver}
\bibfield{author}{\bibinfo{person}{Laurent Granvilliers} {and}
  \bibinfo{person}{Fr{\'e}d{\'e}ric Benhamou}.}
  \bibinfo{year}{2006}\natexlab{}.
\newblock \showarticletitle{Algorithm 852: {{RealPaver}}: An Interval Solver
  Using Constraint Satisfaction Techniques}.
\newblock \bibinfo{journal}{\emph{ACM Trans. Math. Software}}
  \bibinfo{volume}{32}, \bibinfo{number}{1} (\bibinfo{date}{March}
  \bibinfo{year}{2006}), \bibinfo{pages}{138--156}.
\newblock
\showISSN{0098-3500, 1557-7295}
\urldef\tempurl%
\url{https://doi.org/10.1145/1132973.1132980}
\showDOI{\tempurl}


\bibitem[\protect\citeauthoryear{Hesterberg}{Hesterberg}{1995}]%
        {hesterberg1995Weighted}
\bibfield{author}{\bibinfo{person}{Tim Hesterberg}.}
  \bibinfo{year}{1995}\natexlab{}.
\newblock \showarticletitle{Weighted {{Average Importance Sampling}} and
  {{Defensive Mixture Distributions}}}.
\newblock \bibinfo{journal}{\emph{Technometrics}} \bibinfo{volume}{37},
  \bibinfo{number}{2} (\bibinfo{year}{1995}), \bibinfo{pages}{185--194}.
\newblock
\showISSN{0040-1706}
\urldef\tempurl%
\url{https://doi.org/10.2307/1269620}
\showDOI{\tempurl}


\bibitem[\protect\citeauthoryear{Hoffman and Gelman}{Hoffman and
  Gelman}{2014}]%
        {hoffman2014NUTs}
\bibfield{author}{\bibinfo{person}{Matthew~D. Hoffman} {and}
  \bibinfo{person}{Andrew Gelman}.} \bibinfo{year}{2014}\natexlab{}.
\newblock \showarticletitle{The {{No}}-{{U}}-Turn Sampler: Adaptively Setting
  Path Lengths in {{Hamiltonian Monte Carlo}}}.
\newblock \bibinfo{journal}{\emph{The Journal of Machine Learning Research}}
  \bibinfo{volume}{15}, \bibinfo{number}{1} (\bibinfo{date}{Jan.}
  \bibinfo{year}{2014}), \bibinfo{pages}{1593--1623}.
\newblock
\showISSN{1532-4435}


\bibitem[\protect\citeauthoryear{Julian, Lopez, Brush, Owen, and
  Kochenderfer}{Julian et~al\mbox{.}}{2016}]%
        {julian2016Policy}
\bibfield{author}{\bibinfo{person}{Kyle~D. Julian}, \bibinfo{person}{Jessica
  Lopez}, \bibinfo{person}{Jeffrey~S. Brush}, \bibinfo{person}{Michael~P.
  Owen}, {and} \bibinfo{person}{Mykel~J. Kochenderfer}.}
  \bibinfo{year}{2016}\natexlab{}.
\newblock \showarticletitle{Policy Compression for Aircraft Collision Avoidance
  Systems}. In \bibinfo{booktitle}{\emph{2016 {{IEEE}}/{{AIAA}} 35th {{Digital
  Avionics Systems Conference}} ({{DASC}})}}. \bibinfo{publisher}{{IEEE}},
  \bibinfo{pages}{1--10}.
\newblock
\showISBNx{978-1-5090-2523-7}
\urldef\tempurl%
\url{https://doi.org/10.1109/DASC.2016.7778091}
\showDOI{\tempurl}


\bibitem[\protect\citeauthoryear{Katz, Barrett, Dill, Julian, and
  Kochenderfer}{Katz et~al\mbox{.}}{2017}]%
        {katz2017Reluplex}
\bibfield{author}{\bibinfo{person}{Guy Katz}, \bibinfo{person}{Clark Barrett},
  \bibinfo{person}{David~L. Dill}, \bibinfo{person}{Kyle Julian}, {and}
  \bibinfo{person}{Mykel~J. Kochenderfer}.} \bibinfo{year}{2017}\natexlab{}.
\newblock \showarticletitle{Reluplex: {{An Efficient SMT Solver}} for
  {{Verifying Deep Neural Networks}}}. In \bibinfo{booktitle}{\emph{Computer
  {{Aided Verification}}}} \emph{(\bibinfo{series}{Lecture {{Notes}} in
  {{Computer Science}}})}, \bibfield{editor}{\bibinfo{person}{Rupak Majumdar}
  {and} \bibinfo{person}{Viktor Kun{\v c}ak}} (Eds.).
  \bibinfo{publisher}{{Springer International Publishing}},
  \bibinfo{pages}{97--117}.
\newblock
\showISBNx{978-3-319-63387-9}
\urldef\tempurl%
\url{https://doi.org/10.1007/978-3-319-63387-9_5}
\showDOI{\tempurl}


\bibitem[\protect\citeauthoryear{Keramat and Kielbasa}{Keramat and
  Kielbasa}{1998}]%
        {keramat1998study}
\bibfield{author}{\bibinfo{person}{M. Keramat} {and} \bibinfo{person}{R.
  Kielbasa}.} \bibinfo{year}{1998}\natexlab{}.
\newblock \showarticletitle{A Study of Stratified Sampling in Variance
  Reduction Techniques for Parametric Yield Estimation}.
\newblock \bibinfo{journal}{\emph{IEEE Transactions on Circuits and Systems II:
  Analog and Digital Signal Processing}} \bibinfo{volume}{45},
  \bibinfo{number}{5} (\bibinfo{date}{May} \bibinfo{year}{1998}),
  \bibinfo{pages}{575--583}.
\newblock
\showISSN{10577130}
\urldef\tempurl%
\url{https://doi.org/10.1109/82.673639}
\showDOI{\tempurl}


\bibitem[\protect\citeauthoryear{King}{King}{1976}]%
        {king1976Symbolic}
\bibfield{author}{\bibinfo{person}{James~C. King}.}
  \bibinfo{year}{1976}\natexlab{}.
\newblock \showarticletitle{Symbolic Execution and Program Testing}.
\newblock \bibinfo{journal}{\emph{Commun. ACM}} \bibinfo{volume}{19},
  \bibinfo{number}{7} (\bibinfo{date}{July} \bibinfo{year}{1976}),
  \bibinfo{pages}{385--394}.
\newblock
\showISSN{0001-0782, 1557-7317}
\urldef\tempurl%
\url{https://doi.org/10.1145/360248.360252}
\showDOI{\tempurl}


\bibitem[\protect\citeauthoryear{Kingma and Welling}{Kingma and
  Welling}{2014}]%
        {kingma2014AutoEncoding}
\bibfield{author}{\bibinfo{person}{Diederik~P. Kingma} {and}
  \bibinfo{person}{Max Welling}.} \bibinfo{year}{2014}\natexlab{}.
\newblock \showarticletitle{Auto-{{Encoding Variational Bayes}}.}. In
  \bibinfo{booktitle}{\emph{2nd {{International Conference}} on {{Learning
  Representations}}, {{ICLR}} 2014, {{Banff}}, {{AB}}, {{Canada}}, {{April}}
  14-16, 2014, {{Conference Track Proceedings}}}}.
\newblock


\bibitem[\protect\citeauthoryear{Luckow, P{\u a}s{\u a}reanu, Dwyer, Filieri,
  and Visser}{Luckow et~al\mbox{.}}{2014}]%
        {luckow2014Exact}
\bibfield{author}{\bibinfo{person}{Kasper Luckow}, \bibinfo{person}{Corina~S.
  P{\u a}s{\u a}reanu}, \bibinfo{person}{Matthew~B. Dwyer},
  \bibinfo{person}{Antonio Filieri}, {and} \bibinfo{person}{Willem Visser}.}
  \bibinfo{year}{2014}\natexlab{}.
\newblock \showarticletitle{Exact and Approximate Probabilistic Symbolic
  Execution for Nondeterministic Programs}. In
  \bibinfo{booktitle}{\emph{Proceedings of the 29th {{ACM}}/{{IEEE}}
  International Conference on {{Automated}} Software Engineering}}.
  \bibinfo{publisher}{{ACM}}, \bibinfo{pages}{575--586}.
\newblock
\showISBNx{978-1-4503-3013-8}
\urldef\tempurl%
\url{https://doi.org/10.1145/2642937.2643011}
\showDOI{\tempurl}


\bibitem[\protect\citeauthoryear{Luo and Filieri}{Luo and Filieri}{2021}]%
        {luo2021github}
\bibfield{author}{\bibinfo{person}{Yicheng Luo} {and} \bibinfo{person}{Antonio
  Filieri}.} \bibinfo{year}{2021}\natexlab{}.
\newblock \bibinfo{title}{SYMPAIS Implementation}.
\newblock \bibinfo{howpublished}{\url{https://github.com/ethanluoyc/sympais}}.
\newblock


\bibitem[\protect\citeauthoryear{Luo, Filieri, and Zhou}{Luo
  et~al\mbox{.}}{2021}]%
        {luo2021sympais}
\bibfield{author}{\bibinfo{person}{Yicheng Luo}, \bibinfo{person}{Antonio
  Filieri}, {and} \bibinfo{person}{Yuan Zhou}.}
  \bibinfo{year}{2021}\natexlab{}.
\newblock \showarticletitle{Symbolic Parallel Adaptive Importance Sampling for
  Probabilistic Program Analysis}.
\newblock  (\bibinfo{year}{2021}).
\newblock
\showeprint[arxiv]{2010.05050}~[cs.LG]


\bibitem[\protect\citeauthoryear{Malacaria, Khouzani, Pasareanu, Phan, and
  Luckow}{Malacaria et~al\mbox{.}}{2018}]%
        {malacaria2018Symbolic}
\bibfield{author}{\bibinfo{person}{Pasquale Malacaria}, \bibinfo{person}{Mhr
  Khouzani}, \bibinfo{person}{Corina~S. Pasareanu}, \bibinfo{person}{Quoc-Sang
  Phan}, {and} \bibinfo{person}{Kasper Luckow}.}
  \bibinfo{year}{2018}\natexlab{}.
\newblock \showarticletitle{Symbolic {{Side}}-{{Channel Analysis}} for
  {{Probabilistic Programs}}}. In \bibinfo{booktitle}{\emph{2018 {{IEEE}} 31st
  {{Computer Security Foundations Symposium}} ({{CSF}})}}.
  \bibinfo{publisher}{{IEEE}}, \bibinfo{pages}{313--327}.
\newblock
\showISBNx{978-1-5386-6680-7}
\urldef\tempurl%
\url{https://doi.org/10.1109/CSF.2018.00030}
\showDOI{\tempurl}


\bibitem[\protect\citeauthoryear{Martino, Elvira, Luengo, and Corander}{Martino
  et~al\mbox{.}}{2017}]%
        {martino2017Layered}
\bibfield{author}{\bibinfo{person}{L. Martino}, \bibinfo{person}{V. Elvira},
  \bibinfo{person}{D. Luengo}, {and} \bibinfo{person}{J. Corander}.}
  \bibinfo{year}{2017}\natexlab{}.
\newblock \showarticletitle{Layered Adaptive Importance Sampling}.
\newblock \bibinfo{journal}{\emph{Statistics and Computing}}
  \bibinfo{volume}{27}, \bibinfo{number}{3} (\bibinfo{date}{May}
  \bibinfo{year}{2017}), \bibinfo{pages}{599--623}.
\newblock
\showISSN{0960-3174, 1573-1375}
\urldef\tempurl%
\url{https://doi.org/10.1007/s11222-016-9642-5}
\showDOI{\tempurl}


\bibitem[\protect\citeauthoryear{Metropolis, Rosenbluth, Rosenbluth, Teller,
  and Teller}{Metropolis et~al\mbox{.}}{1953}]%
        {metropolis1953Equation}
\bibfield{author}{\bibinfo{person}{Nicholas Metropolis},
  \bibinfo{person}{Arianna~W. Rosenbluth}, \bibinfo{person}{Marshall~N.
  Rosenbluth}, \bibinfo{person}{Augusta~H. Teller}, {and}
  \bibinfo{person}{Edward Teller}.} \bibinfo{year}{1953}\natexlab{}.
\newblock \showarticletitle{Equation of {{State Calculations}} by {{Fast
  Computing Machines}}}.
\newblock \bibinfo{journal}{\emph{The Journal of Chemical Physics}}
  \bibinfo{volume}{21}, \bibinfo{number}{6} (\bibinfo{date}{June}
  \bibinfo{year}{1953}), \bibinfo{pages}{1087--1092}.
\newblock
\showISSN{0021-9606, 1089-7690}
\urldef\tempurl%
\url{https://doi.org/10.1063/1.1699114}
\showDOI{\tempurl}


\bibitem[\protect\citeauthoryear{Neal}{Neal}{2011}]%
        {neal2011MCMC}
\bibfield{author}{\bibinfo{person}{Radford Neal}.}
  \bibinfo{year}{2011}\natexlab{}.
\newblock \showarticletitle{{{MCMC Using Hamiltonian Dynamics}}}.
\newblock In \bibinfo{booktitle}{\emph{Handbook of {{Markov Chain Monte
  Carlo}}}}, \bibfield{editor}{\bibinfo{person}{Steve Brooks},
  \bibinfo{person}{Andrew Gelman}, \bibinfo{person}{Galin Jones}, {and}
  \bibinfo{person}{Xiao-Li Meng}} (Eds.). Vol.~\bibinfo{volume}{20116022}.
  \bibinfo{publisher}{{Chapman and Hall/CRC}}.
\newblock
\showISBNx{978-1-4200-7941-8 978-1-4200-7942-5}
\urldef\tempurl%
\url{https://doi.org/10.1201/b10905-6}
\showDOI{\tempurl}


\bibitem[\protect\citeauthoryear{Nishimura, Dunson, and Lu}{Nishimura
  et~al\mbox{.}}{2020}]%
        {nishimura2020Discontinuous}
\bibfield{author}{\bibinfo{person}{Akihiko Nishimura}, \bibinfo{person}{David~B
  Dunson}, {and} \bibinfo{person}{Jianfeng Lu}.}
  \bibinfo{year}{2020}\natexlab{}.
\newblock \showarticletitle{Discontinuous {{Hamiltonian Monte Carlo}} for
  Discrete Parameters and Discontinuous Likelihoods}.
\newblock \bibinfo{journal}{\emph{Biometrika}} \bibinfo{volume}{107},
  \bibinfo{number}{2} (\bibinfo{date}{June} \bibinfo{year}{2020}),
  \bibinfo{pages}{365--380}.
\newblock
\showISSN{0006-3444, 1464-3510}
\urldef\tempurl%
\url{https://doi.org/10.1093/biomet/asz083}
\showDOI{\tempurl}


\bibitem[\protect\citeauthoryear{Nori, Hur, Rajamani, and Samuel}{Nori
  et~al\mbox{.}}{2014}]%
        {nori2014R2}
\bibfield{author}{\bibinfo{person}{Aditya~V. Nori}, \bibinfo{person}{Chung-Kil
  Hur}, \bibinfo{person}{Sriram~K. Rajamani}, {and} \bibinfo{person}{Selva
  Samuel}.} \bibinfo{year}{2014}\natexlab{}.
\newblock \showarticletitle{R2: An Efficient {{MCMC}} Sampler for Probabilistic
  Programs}. In \bibinfo{booktitle}{\emph{Proceedings of the
  {{Twenty}}-{{Eighth AAAI Conference}} on {{Artificial Intelligence}}}}
  \emph{(\bibinfo{series}{{{AAAI}}'14})}. \bibinfo{publisher}{{AAAI Press}},
  \bibinfo{pages}{2476--2482}.
\newblock


\bibitem[\protect\citeauthoryear{Owen}{Owen}{2013}]%
        {owen2013Monte}
\bibfield{author}{\bibinfo{person}{Art~B. Owen}.}
  \bibinfo{year}{2013}\natexlab{}.
\newblock \bibinfo{booktitle}{\emph{Monte {{Carlo}} Theory, Methods and
  Examples}}.
\newblock
\urldef\tempurl%
\url{https://statweb.stanford.edu/~owen/mc/}
\showURL{%
\tempurl}


\bibitem[\protect\citeauthoryear{P{\u a}s{\u a}reanu and Rungta}{P{\u a}s{\u
  a}reanu and Rungta}{2010}]%
        {pasareanu2010Symbolic}
\bibfield{author}{\bibinfo{person}{Corina~S. P{\u a}s{\u a}reanu} {and}
  \bibinfo{person}{Neha Rungta}.} \bibinfo{year}{2010}\natexlab{}.
\newblock \showarticletitle{Symbolic {{PathFinder}}: Symbolic Execution of
  {{Java}} Bytecode}. In \bibinfo{booktitle}{\emph{Proceedings of the
  {{IEEE}}/{{ACM}} International Conference on {{Automated}} Software
  Engineering - {{ASE}} '10}}. \bibinfo{publisher}{{ACM Press}},
  \bibinfo{pages}{179}.
\newblock
\showISBNx{978-1-4503-0116-9}
\urldef\tempurl%
\url{https://doi.org/10.1145/1858996.1859035}
\showDOI{\tempurl}


\bibitem[\protect\citeauthoryear{Phan, Bang, Pasareanu, Malacaria, and
  Bultan}{Phan et~al\mbox{.}}{2017}]%
        {phan2017Synthesis}
\bibfield{author}{\bibinfo{person}{Quoc-Sang Phan}, \bibinfo{person}{Lucas
  Bang}, \bibinfo{person}{Corina~S. Pasareanu}, \bibinfo{person}{Pasquale
  Malacaria}, {and} \bibinfo{person}{Tevfik Bultan}.}
  \bibinfo{year}{2017}\natexlab{}.
\newblock \showarticletitle{Synthesis of {{Adaptive Side}}-{{Channel
  Attacks}}}. In \bibinfo{booktitle}{\emph{2017 {{IEEE}} 30th {{Computer
  Security Foundations Symposium}} ({{CSF}})}}. \bibinfo{publisher}{{IEEE}},
  \bibinfo{pages}{328--342}.
\newblock
\showISBNx{978-1-5386-3217-8}
\urldef\tempurl%
\url{https://doi.org/10.1109/CSF.2017.8}
\showDOI{\tempurl}


\bibitem[\protect\citeauthoryear{Quarteroni, Sacco, and Saleri}{Quarteroni
  et~al\mbox{.}}{2007}]%
        {quarteroni2007Numerical}
\bibfield{author}{\bibinfo{person}{Alfio Quarteroni}, \bibinfo{person}{Riccardo
  Sacco}, {and} \bibinfo{person}{Fausto Saleri}.}
  \bibinfo{year}{2007}\natexlab{}.
\newblock \bibinfo{booktitle}{\emph{Numerical {{Mathematics}}}
  (\bibinfo{edition}{second} ed.)}.
\newblock \bibinfo{publisher}{{Springer-Verlag}}.
\newblock
\showISBNx{978-3-540-34658-6}
\urldef\tempurl%
\url{https://doi.org/10.1007/b98885}
\showDOI{\tempurl}


\bibitem[\protect\citeauthoryear{Rezende and Mohamed}{Rezende and
  Mohamed}{2015}]%
        {rezende2015Variational}
\bibfield{author}{\bibinfo{person}{Danilo Rezende} {and}
  \bibinfo{person}{Shakir Mohamed}.} \bibinfo{year}{2015}\natexlab{}.
\newblock \showarticletitle{Variational Inference with Normalizing Flows}. In
  \bibinfo{booktitle}{\emph{Proceedings of the 32nd International Conference on
  Machine Learning}} \emph{(\bibinfo{series}{Proceedings of Machine Learning
  Research}, Vol.~\bibinfo{volume}{37})},
  \bibfield{editor}{\bibinfo{person}{Francis Bach} {and} \bibinfo{person}{David
  Blei}} (Eds.). \bibinfo{publisher}{{PMLR}}, \bibinfo{pages}{1530--1538}.
\newblock


\bibitem[\protect\citeauthoryear{Rezende, Mohamed, and Wierstra}{Rezende
  et~al\mbox{.}}{2014}]%
        {rezende2014Stochastic}
\bibfield{author}{\bibinfo{person}{Danilo~Jimenez Rezende},
  \bibinfo{person}{Shakir Mohamed}, {and} \bibinfo{person}{Daan Wierstra}.}
  \bibinfo{year}{2014}\natexlab{}.
\newblock \showarticletitle{Stochastic Backpropagation and Approximate
  Inference in Deep Generative Models}. In
  \bibinfo{booktitle}{\emph{Proceedings of the 31st International Conference on
  Machine Learning}} \emph{(\bibinfo{series}{Proceedings of Machine Learning
  Research}, Vol.~\bibinfo{volume}{32})},
  \bibfield{editor}{\bibinfo{person}{Eric~P. Xing} {and} \bibinfo{person}{Tony
  Jebara}} (Eds.). \bibinfo{publisher}{{PMLR}}, \bibinfo{pages}{1278--1286}.
\newblock


\bibitem[\protect\citeauthoryear{Sankaranarayanan, Chakarov, and
  Gulwani}{Sankaranarayanan et~al\mbox{.}}{2013}]%
        {sankaranarayanan2013Static}
\bibfield{author}{\bibinfo{person}{Sriram Sankaranarayanan},
  \bibinfo{person}{Aleksandar Chakarov}, {and} \bibinfo{person}{Sumit
  Gulwani}.} \bibinfo{year}{2013}\natexlab{}.
\newblock \showarticletitle{Static Analysis for Probabilistic Programs:
  Inferring Whole Program Properties from Finitely Many Paths}. In
  \bibinfo{booktitle}{\emph{Proceedings of the 34th {{ACM SIGPLAN}} Conference
  on {{Programming}} Language Design and Implementation - {{PLDI}} '13}}.
  \bibinfo{publisher}{{ACM Press}}, \bibinfo{pages}{447}.
\newblock
\showISBNx{978-1-4503-2014-6}
\urldef\tempurl%
\url{https://doi.org/10.1145/2491956.2462179}
\showDOI{\tempurl}


\bibitem[\protect\citeauthoryear{Saw, Yang, and Mo}{Saw et~al\mbox{.}}{1984}]%
        {saw1984Chebyshev}
\bibfield{author}{\bibinfo{person}{John~G. Saw}, \bibinfo{person}{Mark~C.K.
  Yang}, {and} \bibinfo{person}{Tse~Chin Mo}.} \bibinfo{year}{1984}\natexlab{}.
\newblock \showarticletitle{Chebyshev {{Inequality}} with {{Estimated Mean}}
  and {{Variance}}}.
\newblock \bibinfo{journal}{\emph{The American Statistician}}
  \bibinfo{volume}{38}, \bibinfo{number}{2} (\bibinfo{date}{May}
  \bibinfo{year}{1984}), \bibinfo{pages}{130--132}.
\newblock
\showISSN{0003-1305, 1537-2731}
\urldef\tempurl%
\url{https://doi.org/10.1080/00031305.1984.10483182}
\showDOI{\tempurl}


\bibitem[\protect\citeauthoryear{Shao, Fienberg, and Olkin}{Shao
  et~al\mbox{.}}{2008}]%
        {shao2008Mathematical}
\bibfield{author}{\bibinfo{person}{Jun Shao}, \bibinfo{person}{S Fienberg},
  {and} \bibinfo{person}{I Olkin}.} \bibinfo{year}{2008}\natexlab{}.
\newblock \bibinfo{booktitle}{\emph{Mathematical {{Statistics}}.}}
\newblock
\showISBNx{978-0-387-21718-5}


\bibitem[\protect\citeauthoryear{Zhou, Yang, Teh, and Rainforth}{Zhou
  et~al\mbox{.}}{2020}]%
        {zhou2020Divide}
\bibfield{author}{\bibinfo{person}{Yuan Zhou}, \bibinfo{person}{Hongseok Yang},
  \bibinfo{person}{Yee~Whye Teh}, {and} \bibinfo{person}{Tom Rainforth}.}
  \bibinfo{year}{2020}\natexlab{}.
\newblock \showarticletitle{Divide, Conquer, and Combine: A New Inference
  Strategy for Probabilistic Programs with Stochastic Support}. In
  \bibinfo{booktitle}{\emph{Proceedings of the 37th International Conference on
  Machine Learning}} \emph{(\bibinfo{series}{Proceedings of Machine Learning
  Research}, Vol.~\bibinfo{volume}{119})},
  \bibfield{editor}{\bibinfo{person}{Hal~Daum{\'e} III} {and}
  \bibinfo{person}{Aarti Singh}} (Eds.). \bibinfo{publisher}{{PMLR}},
  \bibinfo{pages}{11534--11545}.
\newblock


\end{thebibliography}
